\documentclass{article}

 \usepackage[dandb, final]{neurips_2025}

\usepackage[utf8]{inputenc} 
\usepackage[T1]{fontenc}    
\usepackage{hyperref}       
\usepackage{url}            
\usepackage{booktabs}       
\usepackage{amsfonts}       
\usepackage{nicefrac}       
\usepackage{microtype}      
\usepackage{xcolor}         

\usepackage{enumitem}
\usepackage{threeparttable}
\usepackage{graphicx,verbatim}
\usepackage{booktabs} 
\usepackage{multirow}
\usepackage{bbm} 
\usepackage{amssymb}
\usepackage{wrapfig} 
\usepackage{array}
\usepackage{graphicx}
\usepackage{subcaption}
\usepackage{marvosym}
\usepackage{bm}
\usepackage{amsmath}
\usepackage{hyperref}
\usepackage{grffile}
\usepackage{pifont} 
\usepackage{mathtools}
\usepackage{tabularx}
\usepackage{float}
\usepackage{booktabs} 
\usepackage{pifont} 
\usepackage{color}
\usepackage{adjustbox} 
\usepackage[table]{xcolor}
\usepackage{titletoc}
\usepackage{tcolorbox}
\tcbuselibrary{breakable}

\newcommand{\ourmethod}{{\fontfamily{ppl}\selectfont EndoBench}}
\newcommand{\ourinstruct}{{\fontfamily{ppl}\selectfont EndoVQA-Instruct}}

\title{\raisebox{-0.3\height}{\includegraphics[width=1.2cm]{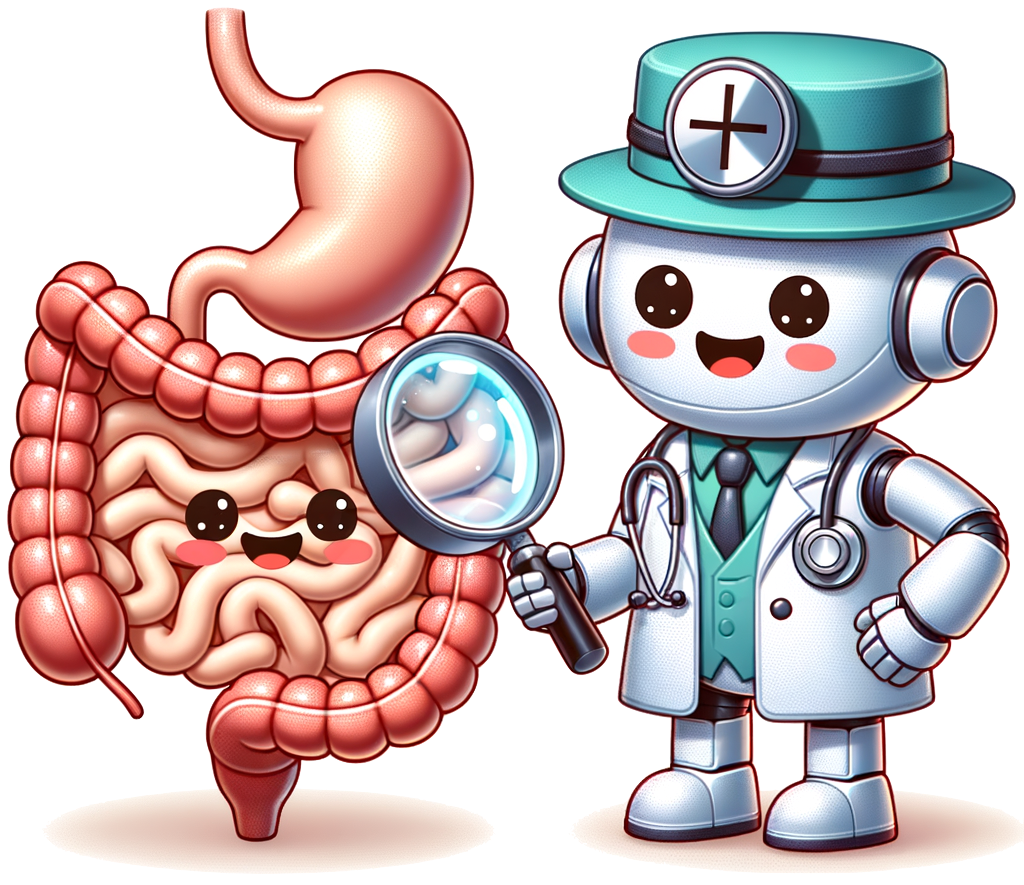}} EndoBench: A Comprehensive Evaluation \\of Multi-Modal Large Language Models \\for Endoscopy Analysis}

\author{Shengyuan Liu$^{1}\thanks{Equal contributions.}$ \quad Boyun Zheng$^{1*}$ \quad Wenting Chen$^{2*}$ \quad Zhihao Peng$^{1}$ \quad 
\\\bf Zhenfei Yin$^{3}$ \quad Jing Shao$^{4}$ \quad Jiancong Hu$^{5}$ \quad Yixuan Yuan$^{1\thanks{Corresponding author (yxyuan@ee.cuhk.edu.hk)}}$ \\
 $^1$Chinese University of Hong Kong \quad
 $^2$City University of Hong Kong  \\
 $^3$University of Oxford \quad
 $^4$ Shanghai AI Laboratory \\
 $^5$ The Sixth Affiliated Hospital, Sun Yat-sen University \\
 \setcounter{footnote}{0}
}

\begin{document}

\maketitle

\begin{abstract}
Endoscopic procedures are essential for diagnosing and treating internal diseases, and multi-modal large language models (MLLMs) are increasingly applied to assist in endoscopy analysis. However, current benchmarks are limited, as they typically cover specific endoscopic scenarios and a small set of clinical tasks, failing to capture the real-world diversity of endoscopic scenarios and the full range of skills needed in clinical workflows. 
To address these issues, we introduce~\ourmethod, the first comprehensive benchmark specifically designed to assess MLLMs across the full spectrum of endoscopic practice with multi-dimensional capacities.~\ourmethod~encompasses 4 distinct endoscopic scenarios, 12 specialized clinical tasks with 12 secondary subtasks, and 5 levels of visual prompting granularities, resulting in 6,832 rigorously validated VQA pairs from 21 diverse datasets. Our multi-dimensional evaluation framework mirrors the clinical workflow—spanning anatomical recognition, lesion analysis, spatial localization, and surgical operations—to holistically gauge the perceptual and diagnostic abilities of MLLMs in realistic scenarios. We benchmark 23 state-of-the-art models, including general-purpose, medical-specialized, and proprietary MLLMs, and establish human clinician performance as a reference standard. Our extensive experiments reveal: (1) proprietary MLLMs outperform open-source and medical-specialized models overall, but still trail human experts; (2) medical-domain supervised fine-tuning substantially boosts task-specific accuracy; and (3) model performance remains sensitive to prompt format and clinical task complexity. EndoBench establishes a new standard for evaluating and advancing MLLMs in endoscopy, highlighting both progress and persistent gaps between current models and expert clinical reasoning. We publicly release our \href{https://cuhk-aim-group.github.io/EndoBench.github.io/}{benchmark and code}.

\end{abstract}

\section{Introduction}


Gastrointestinal and urinary system diseases pose significant global health challenges, where early detection is critical, especially for cancers often diagnosed at advanced stages~\cite{wang2023global,vos2020global,pasechnikov2014gastric}. Endoscopy is a vital diagnostic and therapeutic tool, enabling visualization of internal organs across medical specialties \cite{cao2024wce,colon1,colon2}. As the gold standard for examining internal structures, endoscopy aids in timely pathology detection~\cite{shergill2015role,tringali2015intraductal}. However, the rising demand for endoscopic procedures highlights the need for advanced technologies like artificial intelligence to improve accuracy and efficiency~\cite{esteva2019guide,kroner2021artificial}.

Recent advances in Multi-modal Large Language Models (MLLMs) have produced numerous specialized medical MLLMs dedicated to endoscopy analysis~\cite{colongpt,SurgicalGPT,SurgicalVQLA,SurgicalVQLA_prev,wang2025endochat,hou2024memory,wang2024surgical}. These models enable users to interact through text prompts when analyzing endoscopic images, facilitating various clinical tasks including surgical instrument identification~\cite{SurgicalVQLA,SurgicalVQLA_prev}, lesion detection~\cite{colongpt}, endoscopic image caption~\cite{colongpt,wang2025endochat}, and so on. As these endoscopy-focused MLLMs have developed, there has been a parallel need for robust evaluation frameworks to assess their clinical utility and performance. Existing benchmarks can be categorized into general-purpose~\cite{omnimedvqa,gmaimmbench} and endoscopy-specific evaluations~\cite{kvasir-vqa,SurgicalVQA,SSG-VQA,colongpt}. General benchmarks provide comprehensive assessments across diverse medical data but typically include only limited endoscopic samples covering a narrow range of tasks. Endoscopy-specific benchmarks~\cite{kvasir-vqa,SurgicalVQA,SSG-VQA,colongpt} focus on common procedures such as surgical and colonoscopy, evaluating performance on procedure-specific tasks. Despite these efforts, current benchmarks face significant challenges in assessing whether MLLMs can truly comprehend gastrointestinal endoscopic scenarios with the depth and nuance of clinical professionals.

\begin{table}
  \centering
  \small
  \caption{Comparisons with existing multi-modal endoscopic benchmarks.}
  \label{table:compare}
  \setlength{\tabcolsep}{3pt}
  \begin{threeparttable}
    \begin{tabular}{lllccl}
      \toprule
      Benchmark & Size & Scenario & Task & Granularity & Data Source \\ 
      \midrule
      \tnote{*} OmniMedVQA \cite{omnimedvqa} & 1877 & - & 3 & 1 & 3 Public\\
      \tnote{*} GMAI-MMBench \cite{gmaimmbench} & 3749 & - & 7 & 4 & 16 Public  \\
      Kvasir-VQA \cite{kvasir-vqa} & 6500 & GS,CS & 6 & 1 & 2 Public \\ 
      Surgical-VQA \cite{SurgicalVQA} & 54K & SG & 5 & 1 & 2 Public \\
      SSG-VQA \cite{SSG-VQA} & 960K & SG & 5 & 2 & 3 Public \\      EndoChat~\cite{wang2025endochat} & 396K & SG & 5 & 2 & 3 Public \\
      ColonINST \cite{colongpt} & 300K+ & CS & 4 & 1 & 19 Public \\
      \midrule
      \textbf{\ourinstruct} & 446K+ & GS,CS,CE,SE & 12 & 5 & 20 Public, 1 In-House \\ 
      \textbf{\ourmethod} & 6832 & GS,CS,CE,SE & 12 & 5 & 20 public, 1 In-House \\ 
      \bottomrule
    \end{tabular}
  \end{threeparttable}
\begin{flushleft}
  {\footnotesize
    Abbreviation: GS for Gastroscopy, CS for Colonoscopy, CE for Capsule endoscopy, SE for Surgical endoscopy.
      * The endoscopic data of this benchmark. }
\end{flushleft}
\end{table}

The first challenge lies in the limited scope of existing endoscopic benchmarks, which typically focus on specific scenarios. For instance, Surgical-VQA \cite{SurgicalVQA} and SSG-VQA \cite{SSG-VQA} primarily evaluate the surgical  scenario, while ColonINST \cite{colongpt} concentrates exclusively on colonoscopy. In real clinical settings, however, clinicians always navigate across diverse endoscopic scenarios including Capsule endoscopy (CE), Gastroscopy (GS), Colonoscopy (CS), and Surgical endoscopy (SE). These scenarios differ substantially in their imaging characteristics, anatomical focus, and clinical objectives—ranging from diagnostic screening to interventional procedures—requiring clinicians to possess versatile expertise. The compartmentalized nature of current benchmarks fails to comprehensively assess whether MLLMs can adapt to this multi-modal reality of endoscopic practice. A more holistic, \textit{\textbf{multi-scenario evaluation}} framework that spans all endoscopy scenarios is therefore essential to accurately gauge the clinical utility of these models.

Another challenge is that existing endoscopic VQA benchmarks~\cite{kvasir-vqa,SurgicalVQA,SSG-VQA,colongpt} evaluate only a limited range of tasks, ignoring the multi-dimensional capacities required in clinical practice. While benchmarks like Kvasir-VQA~\cite{kvasir-vqa} focus on basic recognition tasks and ColonINST~\cite{colongpt} emphasizes lesion classification, actual clinical endoscopy follows a structured workflow requiring progressively more sophisticated analysis~\cite{indicators2006quality,kaminski2017performance}. Clinicians always identify organs, recognize anatomical landmarks, detect and classify lesions, quantify findings, precisely localize abnormalities, perform pre-surgical assessments, and execute appropriate interventions. This clinical process demands capabilities spanning from whole-image interpretation to detailed region-level analysis. However, current benchmarks, by focusing on limited tasks rather than this comprehensive spectrum of abilities, inadequately evaluate whether MLLMs can replicate the nuanced expertise that characterizes expert endoscopic assessment. Therefore, a more holistic evaluation framework is needed that assesses model performance across the \textit{\textbf{multi-dimensional capacities}} required in clinical endoscopic examinations.

To address these challenges, we introduce\textbf{~\ourmethod}, a comprehensive endoscopy benchmark designed to evaluate the multi-dimensional capabilities of current multi-modal large language models (MLLMs) in endoscopic image analysis in Fig.~\ref{fig:pipeline}. To the best of our knowledge,~\ourmethod~is the most extensive multi-modal endoscopic benchmark to date, encompassing 4 distinct endoscopic scenarios, 12 specialized endoscopic tasks with 12 secondary subtasks, and 5 levels of visual prompting granularities, as detailed in Table~\ref{table:compare}. For \textit{\textbf{multi-scenario coverage}},~\ourmethod~spans the complete spectrum of endoscopy procedures—from Gastroscopy and Colonoscopy to Capsule endoscopy and Surgical endoscopy. For \textit{\textbf{multi-dimensional capacities evaluation}},~\ourmethod~assesses MLLMs from 12 distinct tasks across 4 major categories, including anatomical structure recognition, lesion analysis and grading, spatial localization and region understanding, and surgical workflow and operation analysis. To thoroughly evaluate fine-grained perceptual capabilities, we implement 5 visual prompting granularities—image-level, box-level, contour-level, multi-box, and multi-contour. Our dataset construction involves collecting 20 public and 1 private endoscopy datasets and standardizing QA pairs, yielding 446,535 VQA pairs comprising our~\ourinstruct~dataset, the current largest endoscopic instruction-tuning collection. From~\ourinstruct, we extract representative pairs that undergo rigorous clinical review, resulting in our final~\ourmethod~dataset of 6,832 clinically validated VQA pairs. For rigorous evaluation, we evaluate 13 open-source general-purpose MLLMs, 5 specialized medical MLLMs, and 5 proprietary general-purpose MLLMs on~\ourmethod. To establish clinical reference standards, we recruit two certified clinicians to answer questions from~\ourmethod. Extensive experiments show that while proprietary MLLMs outperform open-source and specialized models overall, they still lag behind human experts. 

The main contributions of this paper are summarized as follows:
\begin{itemize}
\item We introduce~\ourmethod, the first comprehensive benchmark specifically designed to evaluate MLLMs across the complete spectrum of endoscopy, covering 4 endoscopic scenarios, 12 specialized tasks with 12 secondary subtasks, and 5 levels of visual prompting granularities.

\item We develop the multi-dimensional evaluation framework that mirrors the clinical workflow progression from basic anatomical recognition to advanced surgical intervention, assessing MLLMs' capabilities across the full spectrum of endoscopic analysis skills.

\item We conduct the extensive comparative evaluation of 23 MLLMs (13 open-source general-purpose, 5 medical-specialized, and 5 proprietary models) against human clinician performance, providing insights into current model capabilities.

\end{itemize}

\section{Related Work}
\label{related}

\subsection{Multi-Modal Large Language Models (MLLMs)}
Multi-Modal Large Language Models (MLLMs) are adept at addressing complex multi-modal tasks through large-scale pretraining. Early models like BLIP \cite{BLIP,BLIP2,instructblip} and Flamingo \cite{flamingo} used joint encoders with cross-attention for processing images and text. Later, auto-regressive MLLMs emerged, tokenizing images into visual tokens combined with text tokens for LLM input. Instruction-tuned models like LLaVA \cite{Llava} achieved strong results on vision-language tasks. Recent MLLMs \cite{CogVLM, sharegpt4v, deepseekvl2,januspro,VLIA,internvl2-5}, including QwenVL \cite{qwen2-5vl} and InternVL \cite{internvl2-5}, scaled these architectures, rivaling commercial models like GPT-4o \cite{GPT4o} and Claude-3.7 \cite{Claude3}.
In the medical field, recent studies \cite{Llava-med, medflamingo, QilinMed, huatuogptVision, Meddr, VLIA-M3, Llava-Rad, MedPLIB} have focused on fine-tuning general-purpose models on medical datasets. LLaVA-Med \cite{Llava-med} enhanced LLaVA using PMC-15M \cite{biomedclip} for improved medical VQA performance. HuatuoGPT-Vision \cite{huatuogptVision} created 1.3M medical VQA samples from PubMed, while MedDr \cite{Meddr} employed a retrieval-based approach using InternVL \cite{internvl2-5}. Other works \cite{colongpt, Surgical-lvlm, SurgicalVQA, SurgicalVQLA, GP-VLS} explored MLLMs in endoscopy. ColonGPT \cite{colongpt} aids endoscopists with dialogues, while SurgicalGPT \cite{SurgicalGPT} and Surgical-LVLM \cite{Surgical-lvlm} demonstrated surgical scenario understanding. 


\begin{figure}[t]
  \centering
   \includegraphics[width=\linewidth]{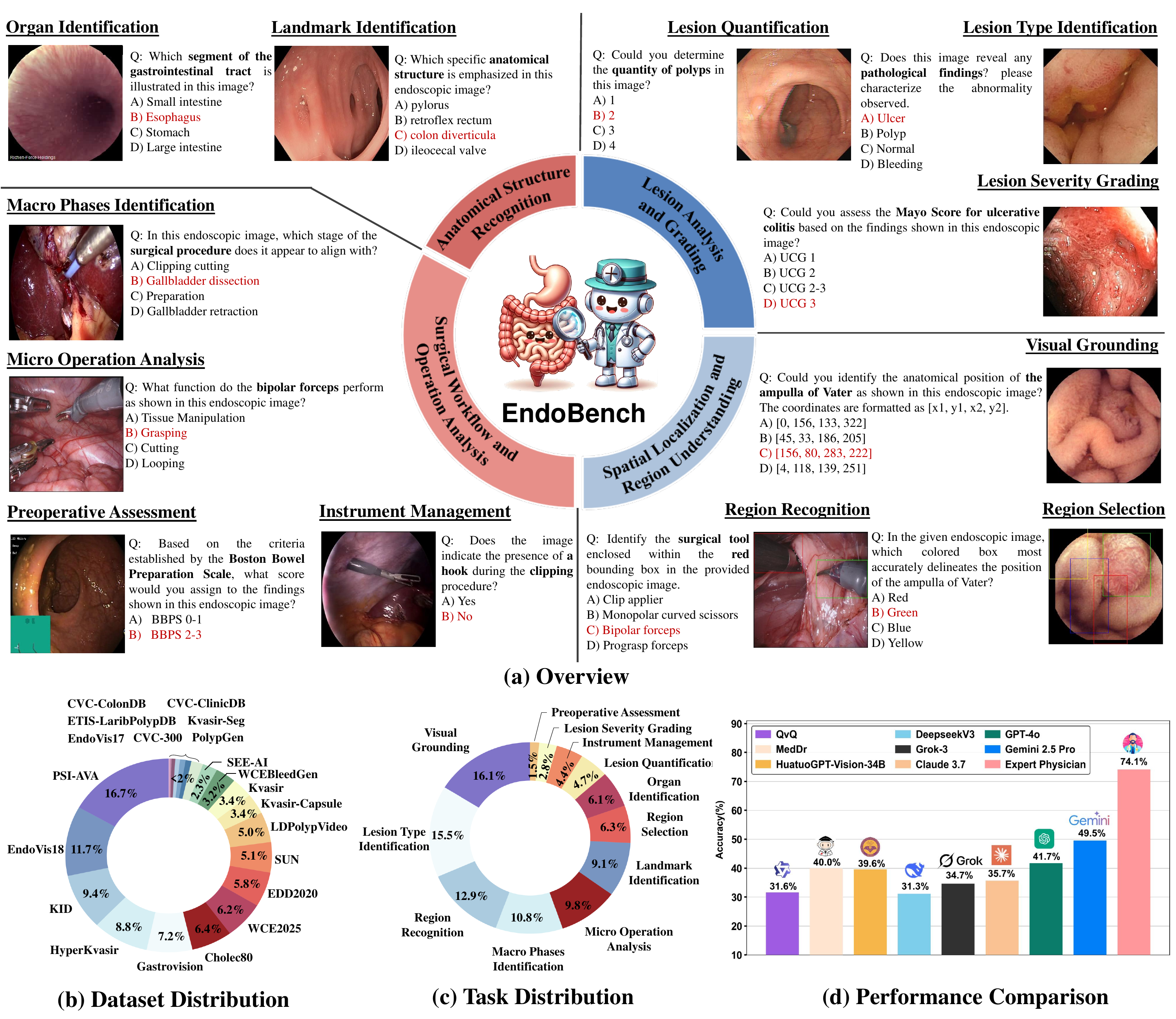}
   \caption{ Overview of our~\textbf{\ourmethod}, the first comprehensive benchmark specifically designed to evaluate MLLMs across the complete spectrum of endoscopy, covering 4 endoscopic scenarios, 12 specialized tasks with 12 secondary subtasks, and 5 levels of visual prompting granularities.}
   \label{fig:pipeline}
\end{figure}    

\subsection{Benchmark for Medical MLLMs}

In the rapidly advancing field of medical MLLMs, numerous benchmarks~\cite{omnimedvqa, Asclepius, Medifusion, gmaimmbench, huatuogptVision, Medxpertqa, PMC-VQA, MedTrinity, Cares} have been developed, offering large and diverse medical VQA datasets to enhance evaluation robustness. GMAI-MMBench \cite{gmaimmbench} incorporated 284 datasets across 38 imaging modalities, 18 clinical tasks, and 18 medical departments. Medifusion \cite{Medifusion} introduced a benchmark with confusing image pairs, requiring distinct answers for identical questions based on subtle image differences. Despite these advancements, endoscopic data remains underrepresented.
In the endoscopic domain, most benchmarks \cite{SurgicalVQA, kvasir-vqa, SSG-VQA, colongpt, PSI-AVA} focus on specific applications. SurgicalVQA \cite{SurgicalVQA} used Cholec80 \cite{Cholec80} and EndoVis-18 \cite{EndoVis18} to evaluate vision-language models in surgery. SSG-VQA \cite{SSG-VQA} tackled laparoscopic tasks like geometric localization and procedure analysis. Kvasir-VQA \cite{kvasir-vqa} added 6,500 question-answer pairs to HyperKvasir \cite{Hyperkvasir} and Kvasir-Instrument \cite{Kvasir-Instrucment}. ColonINST \cite{colongpt} targeted colonoscopy with 303,001 images from 19 datasets. However, these benchmarks lack scenario diversity and task scope, failing to reflect real clinical scenarios. Thus, we propose~\ourmethod, a comprehensive benchmark for evaluating MLLMs across diverse endoscopy applications.


\section{\ourmethod}
\noindent\textbf{Overview.} \ourmethod~is a comprehensive MLLM evaluation framework spanning 4 endoscopy scenarios and 12 clinical tasks with 12 secondary subtasks that mirror the progression of endoscopic examination workflows. Featuring 5 levels of visual prompting granularities to assess region-specific understanding, our \ourmethod~contains 6,832 clinically validated VQA pairs derived from 21 endoscopy datasets. This structure enables precise measurement of MLLMs' clinical perceptual, diagnostic accuracy, and spatial comprehension across diverse endoscopic scenarios.

\subsection{Benchmark Construction}
This section introduces the three main construction steps of~\ourmethod, as shown in Fig.~\ref{fig:pipeline}.

\begin{figure}[t]
  \centering
   \includegraphics[width=\linewidth]{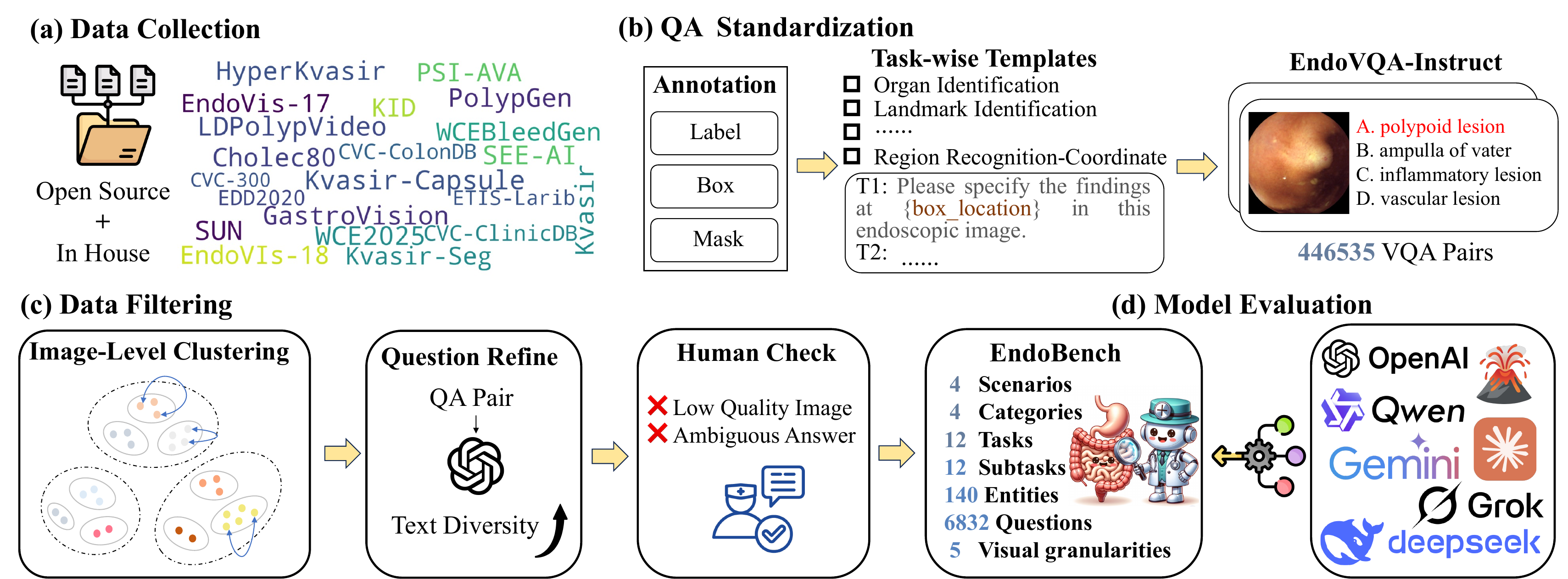}
   \caption{Data construction process of\textbf{~\ourmethod}, consisting of (a) data collection, (b) QA standardization, and (c) data filtering. Finally, we implement (d) model evaluation on~\ourmethod. }
   \label{fig:pipeline}
\end{figure}    

\textbf{Data Collection.} The foundation of our benchmark lies in thorough data collection. Endoscopic images can be classified into two primary categories based on their clinical applications. The first category encompasses diagnostic images used for observation and documentation, including routine upper gastrointestinal endoscopy, lower gastrointestinal colonoscopy, and capsule endoscopy images. The second category consists of therapeutic images used in image-guided minimally invasive procedures, specifically endoscopic surgical images. To ensure comprehensive coverage, on the one hand, we gather 20 public endoscopy datasets from online sources to encompass various endoscopic image types and professional terminology, including Kvasir \cite{Kvasir}, HyperKvasir \cite{Hyperkvasir}, Kvasir-Capsule \cite{Kvasir-capsule}, GastroVision \cite{Gastrovision}, KID \cite{KID}, WCEBleedGen \cite{WCEBleedGen}, SEE-AI \cite{SEE-AI}, Kvasir-Seg \cite{Kvasir-Seg}, CVC-ColonDB \cite{CVC-ColonDB}, ETIS-Larib \cite{ETIS-Larib}, CVC-ClinicDB \cite{CVC-ClinicDB}, CVC-300 \cite{CVC-300}, EDD2020 \cite{EDD2020}, SUN-Database \cite{SUN-database}, LDPolypVideo \cite{LDpolypvideo}, PolypGen \cite{polypgen}, Cholec80 \cite{Cholec80}, EndoVis-17 \cite{endovis17}, EndoVis-18 \cite{EndoVis18}, and PSI-AVA \cite{PSI-AVA}. On the other hand, we further enhance the data diversity by incorporating a private wireless capsule endoscopy image dataset from partner hospitals. All data undergoes the privacy de-identification in accordance with medical ethics requirements. Detailed dataset information is provided in Appendix.

\textbf{QA Standardization.} Following data collection, we standardize the diverse annotation formats across datasets, which include anatomical landmark and pathological lesion labels, as well as structural or lesion annotations in coordinate or image formats (bounding boxes and masks). In collaboration with professional physicians, we develop 12 specialized tasks across 4 major categories and map each dataset's attributes to the appropriate tasks. The 12 specialized tasks include with 12 secondary subtasks. We create 5-8 distinct question templates per task. To facilitate evaluation, each QA pair is supplemented with incorrect answer options, transforming it into a multiple-choice format. Entity labels are naturally conducive to QA pair construction. Question templates are designed based on original categories, with incorrect options randomly selected from attribute nouns of the same type. For spatial comprehension tasks, we standardize coordinate formatting as $[x1, y1, x2, y2]$. To generate plausible distractors for the visual prompts, we create alternative boxes with dimensions similar to the actual bounding box while maintaining overlap below specified thresholds. This process yields 446,535 image-text pairs, comprising our~\ourinstruct~dataset.

\textbf{Data Filtering.} To create a balanced and representative subset for MLLM evaluation, we implement a systematic filtering pipeline on the~\ourinstruct~dataset to obtain our~\ourmethod. We first balance the entity distribution, then employ the DINO-V2 \cite{dinov2} vision foundational model to extract visual embeddings within entities, capturing fine-grained latent representation. Using K-center clustering, we select representative images within each class while maintaining categorical balance and eliminating noise. To enhance dataset diversity and comprehensively evaluate MLLM capabilities, we utilize the GPT-4o-mini API to reformulate questions from the original QA pairs, varying expression styles and syntactic structures while preserving semantic content. This approach enables assessment of MLLMs' adaptability to diverse linguistic representations. Finally, two professional physicians conduct a thorough review of questions and answers, eliminating substandard images and incorrect or ambiguous responses to ensure data quality. Finally, these processes result in our final~\ourmethod~dataset, including 6,832 VQA pairs.

\subsection{Multi-Scenarios Coverage}
Although clinicians regularly navigate diverse endoscopy scenarios in real clinical practice, existing endoscopic benchmarks~\cite{kvasir-vqa,SurgicalVQA,SSG-VQA,colongpt} focus on limited scenarios, making evaluations insufficiently comprehensive. To address this gap,~\ourmethod~encompasses all four major types of endoscopy: Gastroscopy (GS), Colonoscopy (CS), Capsule Endoscopy (CE), and Surgical Endoscopy (SE). The dataset includes 583 samples of GS (8.53\%), 1,848 samples of CS (27.05\%), 1,678 samples of CE (24.56\%), and 2,723 samples of SE (39.86\%), totaling 6,832 samples across all endoscopy types.

\subsection{Multi-Dimensional Capacity} 
Although clinical endoscopy follows a structured workflow requiring progressively more sophisticated analysis~\cite{indicators2006quality,kaminski2017performance}, existing endoscopy benchmarks~\cite{kvasir-vqa,SurgicalVQA,SSG-VQA,colongpt} evaluate only a limited range of tasks, overlooking the multi-dimensional capacities required in clinical practice. To comprehensively evaluate these capacities in MLLMs,~\ourmethod~encompasses 12 clinical tasks with 12 secondary subtasks across 4 major categories for endoscopy analysis. These categories include: (1) \textit{anatomical structure recognition} (organ identification, landmark identification); (2) \textit{lesion analysis and grading} (lesion quantification, lesion type identification, lesion severity grading); (3) \textit{spatial localization and region understanding} (visual grounding, region selection, region recognition); and (4) \textit{surgical workflow and operation analysis} (preoperative assessment, macro phases identification, micro operation analysis, instrument management). The secondary subtasks are detailed in Appendix. The distribution of these tasks is illustrated in Fig.~\ref{fig:pipeline} (b). Moreover, to thoroughly evaluate fine-grained perceptual capabilities of current MLLMs, we implement 5 levels of visual prompting granularities, including image-level, box-level, contour-level, multi-box, and multi-contour. The contour and box are obtained from the original segmentation mask.



\section{Experiments and Analysis}

\subsection{Experiment Setup}
\textbf{Model Evaluation.}  We evaluate 23 MLLMs, comprising 13 open-source models, 5 proprietary models, and 5 medical-domain-specific models. The open-source models, ranging from 3B to 72B parameters, include LLaVA~\cite{Llava}, LLaVA-Next~\cite{llavanext}, CogVLM~\cite{CogVLM}, ShareGPT-4v~\cite{sharegpt4v}, Qwen2.5-VL~\cite{qwen2-5vl}, Janus-Pro~\cite{januspro}, InternVL2.5~\cite{internvl2-5}, and QvQ~\cite{qvq}. Among the proprietary models, we evaluate three reasoning-focused models (Deepseek-V3 with vision, Claude-3.7-Sonnet~\cite{Claude3}, and Gemini-2.5-Pro) and two other MLLMs (GPT-4o~\cite{hurst2024gpt} and Grok-3\cite{grok3}). For medical models, we assess MedDr~\cite{Meddr}, LLaVA-Med~\cite{Llava-med}, HuatuoGPT-Vision~\cite{huatuogptVision}, and the endoscopy-specialized ColonGPT~\cite{colongpt}. 

\textbf{Human Study.} To establish a benchmark for performance, the study includes an evaluation of human clinicians. We randomly select 255 questions from our~\ourmethod~across all the sub-tasks except coordinate-related tasks, due to the precise coordinate format being unsuitable for intuitive human judgment. Each sub-task may include 5, 10, or 15 samples. Two certified clinicians with expertise in endoscopy independently finished the selected questions. Their scores are averaged for each task to provide a reference standard for comparison.

\textbf{Evaluation Metrics.} To evaluate model performance, we measure the accuracy by counting exact matches between predictions and ground-truth answers. For some medical-focused MLLMs that struggle with formatting responses, we use Qwen2.5-VL-72B~\cite{qwen2-5vl} to extract plausible answers for matching. If no valid answer is found, the sample is marked as an error.

\subsection{Experimental Results}

\begin{table}[t]
\small

\caption{Results of different MLLMs on 12 clinical tasks in~\ourmethod. The best-performing model in each category is \textbf{in-bold}, and the second best is \underline{underlined}.}
\label{tab:comparison}
\setlength{\tabcolsep}{1.5pt}
\begin{adjustbox}{width=0.95\textwidth,center}
\begin{tabular}{l|c|cc|ccc|cccc|ccc}
\toprule
 & & \multicolumn{2}{c}{Anatomy} 
 & \multicolumn{3}{c}{Lesion} & \multicolumn{4}{c}{Surgery} & \multicolumn{3}{c}{Spatial} \\ \cmidrule(lr){3-4} \cmidrule(lr){5-7} \cmidrule(lr){8-11}\cmidrule(lr){12-14}
MLLMs & Avg & LI & OI & LQ & LT & LS & PA & MP & MO & IM & VG & RS & RR \\
\midrule
Random & 24.96 & 23.68 & 24.64 & 23.25 & 15.98 & 15.48 & 50.00 & 25.61 & 27.78 & 33.33 & 28.10 & 22.63 & 24.26  \\
Physician & 74.12 & 93.33 & 65.00 & 70.00 & 66.67 & 46.67 & 80.00 & 60.00 & 77.14 & 80.00 & - & 93.33 & 80.00  \\
\rowcolor{orange!20} 
\multicolumn{14}{c}{\textbf{Open-Source MLLMs}} \\ 
Llava-v1.5-7B & 26.62 & 22.24 & 25.84 & 21.64 & 22.16 & 15.79 & 52.00 & 24.25 & 40.24 & 35.69 & 22.87 & 25.64 & 28.99 \\
Llava-v1.5-13B & 24.99 & 24.80 & 19.14 & 19.75 & 12.89 & 18.89 & 56.00 & 22.36 & 46.55 & 34.68 & 19.73 & 26.56 & 22.58  \\
Llava-llama3-8B & 24.75 & 22.56 & 21.77 & 22.40 & 18.56 & 18.27 & 51.00 & 23.71 & 42.49 & 37.04 & 18.68 & 25.40 & 20.82  \\
Llava-Next-Llama3-8B & 25.10 & 27.04 & 23.21 & 17.01 & 18.04 & 17.03 & 51.00 & 24.53 & 44.89 & 39.73 & 18.24 & 26.79 & 22.27  \\
CogVLM-Chat-7B & 27.58 & 24.64 & 25.60 & 21.08 & 20.10 & 18.58 & 50.00 & 23.71 & 41.44 & 34.68 & 26.76 & 24.94 & 31.27  \\
ShareGPT-4v & 18.35 & 11.52 & 18.42 & 19.19 & 1.55 & 14.24 & 0.00 & 16.26 & 31.38 & 21.55 & 11.51 & 12.70 & 25.02  \\
Qwen2.5VL-3B-Instruct & 25.01 & 21.60 & 23.44 & 20.70 & 15.98 & 16.72 & 44.00 & 26.02 & 42.34 & 50.17 & 17.49 & 25.87 & 21.05  \\
Qwen2.5VL-7B-Instruct & 27.63 & 22.24 & 25.84 & 21.64 & 22.16 & 15.79 & 55.00 & 24.25 & 40.24 & 57.91 & 22.87 & 25.64 & 28.99  \\
Qwen2.5VL-72B-Instruct & 27.25 & 28.48 & 20.10 & 22.21 & 12.37 & 14.55 & 53.00 & 27.51 & 48.05 & 50.17 & 22.87 & 26.10 & 23.11  \\
Janus-Pro-7B & 28.81 & 25.28 & 23.68 & 22.31 & 13.40 & 17.03 & 50.00 & 27.37 & 47.90 & 45.45 & 25.56 & 25.87 & 30.89  \\
InternVL2.5-8B & 27.96 & 23.20 & 20.10 & 19.09 & 8.76 & 17.96 & 54.00 & 26.83 & 49.25 & 45.79 & 16.74 & 26.10 & 35.32\\
InternVL2.5-38B & 30.09 & 28.48 & 31.82 & 22.02 & 13.40 & 19.50 & 57.00 & 27.64 & 47.60 & 49.83 & 19.13 & 28.18 & 34.10 \\
QvQ-72B & 31.62 & 22.08 & 15.31 & 30.91 & 22.68 & 18.89 & 53.00 & 28.86 & 49.85 & 53.87 & 28.85 & 37.88 & 31.35  \\
\rowcolor{green!20} 
\multicolumn{14}{c}{\textbf{Open-Source Medical-Domain MLLMs}} \\ 
MedDr-80B & 39.96 & \textbf{56.00} & \textbf{43.06} & 36.96 & 21.65 & 17.65 & 52.00 & 28.05 & 57.51 & 48.48 & \underline{45.14} & 47.58 & 31.73 \\
Llava-Med-7B & 24.71 & 41.44 & 26.79 & 15.79 & 24.23 & 8.67 & 47.00 & 17.61 & 24.93 & 25.26 & 24.36 & 37.88 & 25.17 \\
HuatuoGPT-Vision-7B & 35.57 & 34.88 & 39.47 & 37.43 & 23.20 & 21.05 & 45.00 & 31.03 & 49.85 & 49.49 & 24.07 & 46.42 & 32.26 \\
HuatuoGPT-Vision-34B & 39.58 & 36.64 & 33.25 & 34.97 & 19.59 & 21.98 & 86.00 & 35.77 & 55.26 & 59.26 & 31.69 & \underline{60.51} & \underline{37.30}  \\
ColonGPT & 15.60 & 30.40 & 11.00 & 27.69 & 12.37 & 0.00 & \textbf{95.00} & 5.42 & 1.65 & 15.83 & 2.99 & 4.62 & 21.36  \\
\rowcolor{pink!20} 
\multicolumn{14}{c}{\textbf{Proprietary MLLMs}} \\ 
Deepseek-V3 & 31.13 & 24.96 & 21.53 & 29.96 & 21.65 & 20.43 & 46.00 & 27.91 & 52.10 & 54.55 & 21.38 & 38.57 & 29.37 \\
Grok-3 & 34.66 & 32.16 & 26.08 & 33.65 & 16.49 & 22.29 & 56.00 & 27.38 & 49.25 & 53.87 & 20.93 & 54.73 & 36.23\\
Claude-3.7-Sonnet & 35.67 & 34.56 & 26.07 & 36.01 & \underline{29.90} & \textbf{25.39} & 44.00 & 27.78 & 53.76 & 48.48 & 27.65 & 51.27 & 33.03 \\
GPT-4o & \underline{41.69} & \underline{44.16} & 33.73 & \textbf{42.25} & \textbf{39.69} & 24.15 & \underline{92.00} & \underline{41.19} & \underline{59.16} & \textbf{63.63} & 27.06 & 41.80 & 37.22 \\
Gemini-2.5-Pro & \textbf{49.53} & \underline{44.16} & \underline{39.71} & \underline{41.97} & 29.38 & \underline{24.46} & 90.00 & \textbf{46.21} & \textbf{67.87} & \underline{62.96} & \textbf{50.52} & \textbf{73.21} & \textbf{48.59} \\
\bottomrule
\end{tabular}
\end{adjustbox}
\begin{flushleft}
  {\footnotesize
   Abbreviation: Anatomy for Anatomical Structure Recognition, Lesion for Lesion Analysis and Grading, Surgery for Surgical Workflow and Operation Analysis, Spatial for Spatial localization and region understanding. The abbreviations for the corresponding tasks are defined in Appendix.
   }
\end{flushleft}
\end{table}

\begin{figure}[t]
  \centering
   \includegraphics[width=\linewidth]{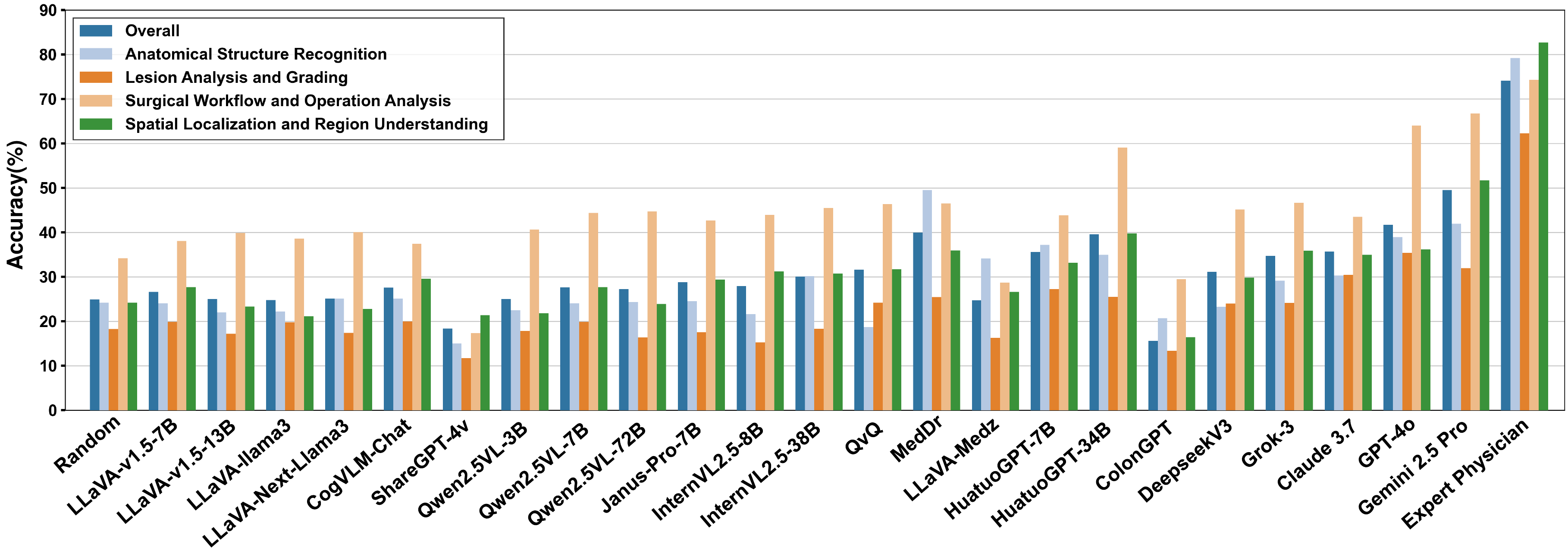}
   \caption{Performance comparison across 4 major categories in~\ourmethod~among existing MLLMs. }
   \label{fig:major_category}
\end{figure}

\textbf{Results across Capacities.}
Table~\ref{tab:comparison} summarizes the performance of various MLLMs across 12 clinical tasks. Proprietary models dominate, with Gemini-2.5-Pro achieving the highest average score (49.53\%) and excelling in 7 tasks, though still far from human physician performance (74.12\%). GPT-4o follows with strong results (41.69\%), especially in lesion analysis. Medical-domain models like MedDr-80B and HuatuoGPT-Vision-34B perform well in specific tasks but lag overall. ColonGPT shows extreme variability, excelling in Preoperative Assessment (95\%, surpassing physicians at 80\%) but underperforming elsewhere. Open-source models generally trail, though QvQ-72B (31.62\%) and InternVL2.5-38B (30.09\%) show promise. These findings reveal proprietary MLLMs' overall superiority but underscore the gap between MLLMs and human expertise. Fig.~\ref{fig:major_category} illustrates the performance of MLLMs across four major categories. Most models consistently achieve their highest performance on surgery-related tasks, while struggling most with lesion-relevant tasks, highlighting the varying difficulty levels these categories present for current MLLMs.


\textbf{Results across Scenarios.}
Table~\ref{tab:comparison_scene} compares performance of different MLLMs across endoscopy scenarios. Proprietary MLLMs, particularly Gemini-2.5-Pro (52.39\%), outperform all other models across clinical tasks and visual prompts, with GPT-4o (42.87\%) following as a strong competitor. Medical-domain models, like HuatuoGPT-Vision-34B (41.55\%), show potential in specific tasks but lack consistency. Open-source models generally underperform, though QvQ-72B (33.01\%) and InternVL2.5-38B (32.36\%) demonstrate some promise. ColonGPT is a notable outlier, excelling in Preoperative Assessment (95\%) but performing poorly overall (10.47\%). Despite advancements, all MLLMs lag human physicians, who achieve a superior average score of 76.64\%, emphasizing the need for further development to bridge this performance gap.

\begin{table}[]
\caption{Results of different MLLMs on 4 different endoscopy scenarios and 4 different visual prompts in~\ourmethod. The best-performing model in each category is \textbf{in-bold}, and the second best is \underline{underlined}.}
\label{tab:comparison_scene}
\setlength{\tabcolsep}{1.5pt}
\begin{adjustbox}{width=0.8\textwidth,center}
\footnotesize
\begin{tabular}{l|ccccc|ccccc}
\toprule
\multicolumn{1}{c}{\multirow{2}{*}{MLLMs}} & \multicolumn{5}{c}{Endoscopy Scenarios} & \multicolumn{5}{c}{Viusal Prompt} \\
\cmidrule(lr){2-6} \cmidrule(lr){7-11}
\multicolumn{1}{c}{} & \multicolumn{1}{c}{Avg} & \multicolumn{1}{c}{GS} & \multicolumn{1}{c}{CS} & \multicolumn{1}{c}{CE} & \multicolumn{1}{c}{SE} & \multicolumn{1}{c}{Avg} & \multicolumn{1}{c}{Box} & \multicolumn{1}{c}{Cont} & \multicolumn{1}{c}{Mul} & \multicolumn{1}{c}{Coor} \\
\midrule
Random & 25.58 & 24.01 & 23.41 & 25.48 & 26.35 & 23.61 & 26.10 & 19.23 & 25.10 & 24.02 \\
Physician & 76.64 & 80.00 & 81.94 & 78.26 & 74.59 & 80.00 & 86.67 & 73.33 & 80.00 & - \\
\rowcolor{orange!20} 
\multicolumn{11}{c}{\textbf{Open-Source MLLMs}} \\ 
Llava-v1.5-7B & 27.30 & 15.09 & 28.50 & 22.35 & 30.49 & 29.11& 34.18 & 32.97 & 24.33 & 24.94 \\
Llava-v1.5-13B & 26.22 & 15.61 & 22.14 & 20.86 & 30.23 & 23.57 & 19.17 & 23.08 & 31.94 & 20.09 \\
Llava-llama3-8B & 26.87 & 27.79 & 19.86 & 22.47 & 28.96 & 20.74 & 19.17 & 17.58 & 24.71 & 21.48 \\
Llava-Next-Llama3-8B & 27.02 & 23.33 & 21.35 & 22.41 & 29.82 & 22.96 & 22.63 & 21.98 & 29.66 & 17.55 \\
CogVLM-Chat-7B & 29.23 & 18.52 & 26.32 & 23.18 & 33.17 & 30.96 & 31.87 & 29.67 & 30.42 & 31.87 \\
ShareGPT-4v-7B & 19.11 & 12.52 & 18.17 & 15.97 & 21.24 & 25.06 & 29.10 & 29.12 & 19.39 & 22.63 \\
Qwen2.5VL-3B-Instruct & 27.81 & 29.85 & 18.17 & 22.41 & 30.41 & 22.05& 18.48 & 24.18 & 26.62 & 18.94 \\
Qwen2.5VL-7B-Instruct & 20.95 & 14.58 & 21.77 & 20.02 & 22.25 & 29.11 & 34.18 & 32.97 & 24.33 & 24.94 \\
Qwen2.5VL-72B-Instruct & 29.57 & 25.56 & 22.99 & 23.66 & 32.87 & 23.76 & 21.02 & 24.73 & 27.38 & 21.94 \\
Janus-Pro-7B & 31.12 & 26.93 & 23.99 & 24.20 & 34.77 & 30.83 & 32.33 & 26.92 & \underline{36.12} & 27.94 \\
InternVL2.5-8B & 29.94 & 20.75 & 27.44 & 21.39 & 33.99 & 34.99 & 40.42 & 34.07 & 33.84 & 31.64 \\
InternVL2.5-38B & 32.36 & 28.99 & 25.85 & 26.64 & 35.48 & 33.48 & 38.57 & 31.32 & 31.94 & 32.10 \\
QvQ-72B & 33.01 & 31.73 & 29.93 & 25.03 & 35.48 & 30.88 & 34.41 & 31.32 & 26.62 & 31.18 \\
\rowcolor{green!20} 
\multicolumn{11}{c}{\textbf{Open-Source Medical-Domain MLLMs}} \\ 
MedDr-80B & 40.92 & 51.46 & 37.76 & 38.50 & 39.92 & 31.73 & 33.03 & 34.62 & 27.38 & 31.87 \\
Llava-Med-7B & 25.11 & 35.33 & 24.10 & 23.06 & 23.67 & 24.71 & 25.64 & 23.08 & 23.57 & 26.56 \\
HuatuoGPT-Vision-7B & 36.04 & 36.88 & 34.32 & \underline{35.22} & 36.38 & 32.40 & 32.56 & 35.71 & 28.52 & 32.79 \\
HuatuoGPT-Vision-34B & 41.55 & 45.80 & 38.14 & 33.61 & 42.97 & \underline{37.20} & 39.49 & \underline{37.91} & 35.36 & 36.03 \\
ColonGPT & 10.47 & 9.61 & 33.37 & 16.51 & 4.85 & 21.55 & 24.71 & 34.07 & 4.56 & 22.86 \\
\rowcolor{pink!20} 
\multicolumn{11}{c}{\textbf{Proprietary MLLMs}} \\ 
Deepseek-V3 & 32.34 & 27.79 & 30.46 & 27.53 & 34.59 & 29.86 & 31.18 & 34.07 & 27.38 & 26.79 \\
Grok-3 & 35.37 & 34.31 & 31.30 & 36.00 & 36.27 & 34.86 & \underline{41.57} & 32.97 & 27.00 & 37.88 \\
Claude-3.7-Sonnet & 36.26 & 37.91 & 35.01 & 34.15 & 36.60 & 33.12 & 30.48 & 35.71 & 30.04 & 36.26 \\
GPT-4o & \underline{42.87} & \underline{45.97} & \underline{43.54} & 34.86 & \underline{43.72} & 36.78 & 32.79 & 35.71 & 34.98 & \underline{43.65} \\
Gemini-2.5-Pro & \textbf{52.39} & \textbf{57.29} & \textbf{44.60} & \textbf{44.22} & \textbf{54.60} & \textbf{47.39} & \textbf{49.19} & \textbf{38.46} & \textbf{51.33} & \textbf{50.58} \\
\bottomrule
\end{tabular}
\end{adjustbox}

\begin{flushleft}
  {\footnotesize
  Abbreviation: Cont for Contour, Mul for Multi-region, Coor for Coordinate.
  }
  \end{flushleft}
\end{table}

\subsection{Discussion}

From the above results, four key insights have been deduced as follows:

\textbf{1) Endoscopy remains a challenging domain for MLLMs, with significant gaps between models and human expertise.} Human experts achieve an average accuracy of 74.12\% in endoscopy tasks, while the top-performing model, Gemini-2.5-Pro, reaches only 49.53\%—a gap of roughly 25\%. This highlights the inherent difficulty of endoscopy, which demands both precise visual interpretation and specialized medical knowledge. Proprietary models consistently outperform open-source models overall, yet open-source models show a surprising edge in surgical scenarios, where their accuracy improves markedly compared to random baselines. In contrast, for non-surgical tasks like landmark and organ identification, open-source models perform no better than random guessing. This disparity suggests that while open-source models can leverage structured contexts, they falter in knowledge-intensive tasks, pointing to a need for enhanced domain-specific capabilities.

\begin{figure}[t]
  \centering
   \includegraphics[width=\linewidth]{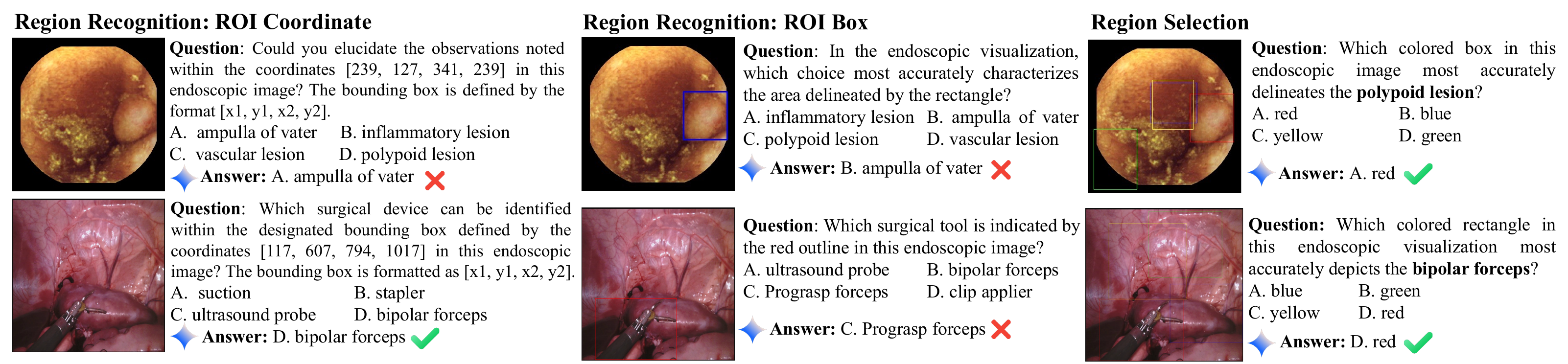}
   \caption{Case study reveals that model performance varies with different visual prompt formats.}
   \label{fig:visual_prompt}
\end{figure} 

\textbf{2) Medical domain-specific Supervised Fine-Tuning markedly boosts model performance.} Medical models that underwent domain-specific supervised fine-tuning, such as MedDr and HuatuoGPT-Vision-34B, perform exceptionally well in tasks like landmark identification and organ recognition, even outperforming all proprietary models. This indicates that domain pretraining effectively equips models with essential medical knowledge, enhancing their competitiveness in specialized tasks. However, some medical models exhibit limitations in instruction-following capabilities and suffer from overfitting, which restricts their performance in broader application scenarios. This suggests that while conducting domain-specific training, greater attention should be paid to balancing model generalization and task adaptability.
   
\textbf{3) Model performance varies with visual prompt formats, exposing a gap between visual perception and medical comprehension.} The ability of models to understand spatial information varies significantly based on how visual prompts are formatted. To explore this, we test the same images across 3 tasks with different visual prompts, and the results in Table \ref{tab:comparison} and Table \ref{tab:comparison_scene} reveal that most models, especially proprietary ones, excelled in the Region Selection task, indicating strong visual comprehension in distinguishing between regions. However, they struggle to accurately classify lesion types within those regions, pointing to a lack of medical knowledge as the main source of errors rather than poor visual processing. A case study is shown in Fig. \ref{fig:visual_prompt}, and it suggests that while models can spatially differentiate key areas, their interpretation hinges on both the prompt format and their insufficient medical knowledge. 

\begin{figure}[t]
  \centering
   \includegraphics[width=\linewidth]{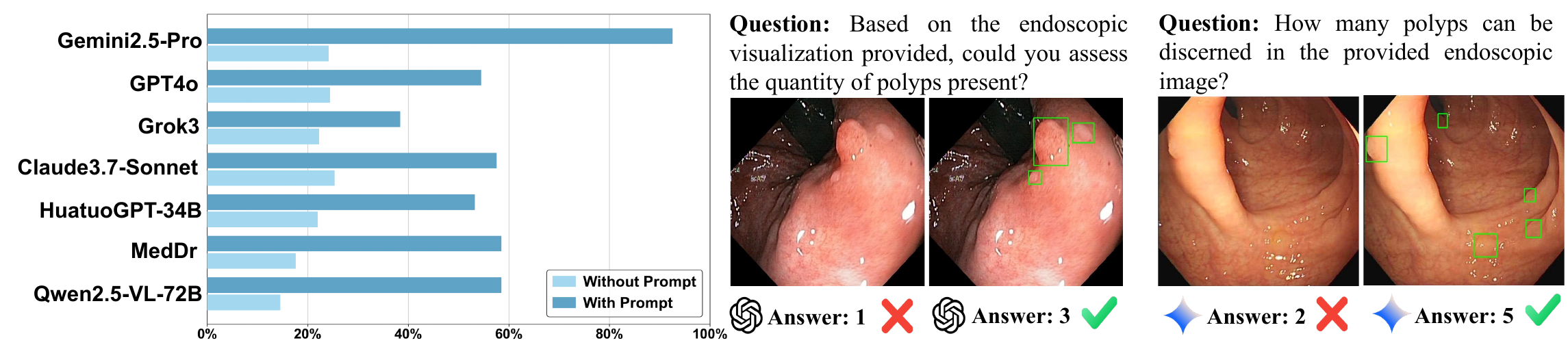}
   \caption{The influence of visual prompt in lesion quantification task among different MLLMs.}
   \label{fig:polyp_case}
\end{figure}

\textbf{4) Polyp counting exposes dual challenges in lesion identification and numerical reasoning.} Initial testing reveals severe limitations in this task, with no model achieving above 30\% accuracy. To better understand these performance issues, we add bounding boxes as visual prompts (Fig. \ref{fig:polyp_case}), which dramatically improve accuracy across all models. Most notably, Gemini-2.5-Pro achieves 92.57\% from 24.46\% with this new prompting approach. This improvement suggests that while Gemini possesses robust spatial recognition and counting abilities, the primary challenge for models lies not in computational or spatial reasoning but in lesion identification. Our findings highlight the importance of incorporating domain-specific medical knowledge into MLLMs to enhance their performance in tasks that combine visual analysis with clinical expertise.



\section{Conclusion}
We introduce~\ourmethod, the most comprehensive benchmark to date for evaluating multi-modal large language models in endoscopic image analysis. Our results show that while proprietary and domain-adapted MLLMs outperform open-source models in many tasks, all models still fall significantly short of human clinician performance—especially in complex, nuanced clinical scenarios.~\ourmethod~exposes key limitations in current MLLMs’ clinical dignaosis and spatial understanding, highlighting the need for further research in domain adaptation and prompt design. We hope EndoBench will serve as a valuable resource for advancing clinically relevant AI in endoscopy.



\bibliographystyle{nips}
\bibliography{ref}

\begin{thebibliography}{100}

\bibitem{wang2023global}
Wang, Y., Y.~Huang, R.~C. Chase, et~al.
\newblock Global burden of digestive diseases: a systematic analysis of the global burden of diseases study, 1990 to 2019.
\newblock \emph{Gastroenterology}, 165(3):773--783, 2023.

\bibitem{vos2020global}
Vos, T., S.~S. Lim, C.~Abbafati, et~al.
\newblock Global burden of 369 diseases and injuries in 204 countries and territories, 1990--2019: a systematic analysis for the global burden of disease study 2019.
\newblock \emph{The lancet}, 396(10258):1204--1222, 2020.

\bibitem{pasechnikov2014gastric}
Pasechnikov, V., S.~Chukov, E.~Fedorov, et~al.
\newblock Gastric cancer: prevention, screening and early diagnosis.
\newblock \emph{World journal of gastroenterology: WJG}, 20(38):13842, 2014.

\bibitem{cao2024wce}
Cao, Q., R.~Deng, Y.~Pan, et~al.
\newblock Robotic wireless capsule endoscopy: recent advances and upcoming technologies.
\newblock \emph{Nature Communications}, 15(1):4597, 2024.

\bibitem{colon1}
Eng, C., T.~Yoshino, E.~Ru{\'\i}z-Garc{\'\i}a, et~al.
\newblock Colorectal cancer.
\newblock \emph{Lancet}, 394(10207):1467--1480, 2024.

\bibitem{colon2}
Wallace, M.~B., P.~Sharma, P.~Bhandari, et~al.
\newblock Impact of artificial intelligence on miss rate of colorectal neoplasia.
\newblock \emph{Gastroenterology}, 163(1):295--304, 2022.

\bibitem{shergill2015role}
Shergill, A.~K., J.~R. Lightdale, D.~H. Bruining, et~al.
\newblock The role of endoscopy in inflammatory bowel disease.
\newblock \emph{Gastrointestinal endoscopy}, 81(5):1101--1121, 2015.

\bibitem{tringali2015intraductal}
Tringali, A., A.~Lemmers, V.~Meves, et~al.
\newblock Intraductal biliopancreatic imaging: European society of gastrointestinal endoscopy (esge) technology review.
\newblock \emph{Endoscopy}, 47(08):739--753, 2015.

\bibitem{esteva2019guide}
Esteva, A., A.~Robicquet, B.~Ramsundar, et~al.
\newblock A guide to deep learning in healthcare.
\newblock \emph{Nature medicine}, 25(1):24--29, 2019.

\bibitem{kroner2021artificial}
Kr{\"o}ner, P.~T., M.~M. Engels, B.~S. Glicksberg, et~al.
\newblock Artificial intelligence in gastroenterology: A state-of-the-art review.
\newblock \emph{World journal of gastroenterology}, 27(40):6794, 2021.

\bibitem{colongpt}
Ji, G.-P., J.~Liu, P.~Xu, et~al.
\newblock Frontiers in intelligent colonoscopy.
\newblock \emph{arXiv preprint arXiv:2410.17241}, 2024.

\bibitem{SurgicalGPT}
Seenivasan, L., M.~Islam, G.~Kannan, et~al.
\newblock Surgicalgpt: end-to-end language-vision gpt for visual question answering in surgery.
\newblock In \emph{MICCAI}, pages 281--290. Springer, 2023.

\bibitem{SurgicalVQLA}
Bai, L., G.~Wang, M.~Islam, et~al.
\newblock Surgical-vqla++: Adversarial contrastive learning for calibrated robust visual question-localized answering in robotic surgery.
\newblock \emph{Information Fusion}, 113:102602, 2025.

\bibitem{SurgicalVQLA_prev}
Bai, L., M.~Islam, L.~Seenivasan, et~al.
\newblock Surgical-vqla: Transformer with gated vision-language embedding for visual question localized-answering in robotic surgery.
\newblock In \emph{2023 IEEE International Conference on Robotics and Automation (ICRA)}, pages 6859--6865. IEEE, 2023.

\bibitem{wang2025endochat}
Wang, G., L.~Bai, J.~Wang, et~al.
\newblock Endochat: Grounded multimodal large language model for endoscopic surgery.
\newblock \emph{arXiv preprint arXiv:2501.11347}, 2025.

\bibitem{hou2024memory}
Hou, W., Y.~Cheng, K.~Xu, et~al.
\newblock Memory-augmented multimodal llms for surgical vqa via self-contained inquiry.
\newblock \emph{arXiv preprint arXiv:2411.10937}, 2024.

\bibitem{wang2024surgical}
Wang, G., L.~Bai, W.~J. Nah, et~al.
\newblock Surgical-lvlm: Learning to adapt large vision-language model for grounded visual question answering in robotic surgery.
\newblock \emph{arXiv preprint arXiv:2405.10948}, 2024.

\bibitem{omnimedvqa}
Hu, Y., T.~Li, Q.~Lu, et~al.
\newblock Omnimedvqa: A new large-scale comprehensive evaluation benchmark for medical lvlm.
\newblock In \emph{CVPR}, pages 22170--22183. 2024.

\bibitem{gmaimmbench}
Ye, J., G.~Wang, Y.~Li, et~al.
\newblock Gmai-mmbench: A comprehensive multimodal evaluation benchmark towards general medical ai.
\newblock \emph{NeurIPS}, 37:94327--94427, 2024.

\bibitem{kvasir-vqa}
Gautam, S., A.~M. Stor{\aa}s, C.~Midoglu, et~al.
\newblock Kvasir-vqa: A text-image pair gi tract dataset.
\newblock In \emph{Proceedings of the First International Workshop on Vision-Language Models for Biomedical Applications}, pages 3--12. 2024.

\bibitem{SurgicalVQA}
Seenivasan, L., M.~Islam, A.~K. Krishna, et~al.
\newblock Surgical-vqa: Visual question answering in surgical scenes using transformer.
\newblock In \emph{MICCAI}, pages 33--43. Springer, 2022.

\bibitem{SSG-VQA}
Yuan, K., M.~Kattel, J.~L. Lavanchy, et~al.
\newblock Advancing surgical vqa with scene graph knowledge.
\newblock \emph{International Journal of Computer Assisted Radiology and Surgery}, pages 1--9, 2024.

\bibitem{indicators2006quality}
INDICATORS, S.~Q.
\newblock Quality indicators for colonoscopy.
\newblock \emph{Am J Gastroenterol}, 101:873--885, 2006.

\bibitem{kaminski2017performance}
Kaminski, M.~F., S.~Thomas-Gibson, M.~Bugajski, et~al.
\newblock Performance measures for lower gastrointestinal endoscopy: a european society of gastrointestinal endoscopy (esge) quality improvement initiative.
\newblock \emph{Endoscopy}, 49(04):378--397, 2017.

\bibitem{BLIP}
Li, J., D.~Li, C.~Xiong, et~al.
\newblock Blip: Bootstrapping language-image pre-training for unified vision-language understanding and generation.
\newblock In \emph{ICML}, pages 12888--12900. PMLR, 2022.

\bibitem{BLIP2}
Li, J., D.~Li, S.~Savarese, et~al.
\newblock Blip-2: Bootstrapping language-image pre-training with frozen image encoders and large language models.
\newblock In \emph{ICML}, pages 19730--19742. PMLR, 2023.

\bibitem{instructblip}
Dai, W., J.~Li, D.~Li, et~al.
\newblock Instructblip: Towards general-purpose vision-language models with instruction tuning, 2023.

\bibitem{flamingo}
Alayrac, J.-B., J.~Donahue, P.~Luc, et~al.
\newblock Flamingo: a visual language model for few-shot learning.
\newblock \emph{NeurIPS}, 35:23716--23736, 2022.

\bibitem{Llava}
Liu, H., C.~Li, Q.~Wu, et~al.
\newblock Visual instruction tuning.
\newblock \emph{NeurIPS}, 36:34892--34916, 2023.

\bibitem{CogVLM}
Wang, W., Q.~Lv, W.~Yu, et~al.
\newblock Cogvlm: Visual expert for pretrained language models.
\newblock \emph{NeurIPS}, 37:121475--121499, 2024.

\bibitem{sharegpt4v}
Chen, L., J.~Li, X.~Dong, et~al.
\newblock Sharegpt4v: Improving large multi-modal models with better captions.
\newblock In \emph{European Conference on Computer Vision}, pages 370--387. Springer, 2024.

\bibitem{deepseekvl2}
Wu, Z., X.~Chen, Z.~Pan, et~al.
\newblock Deepseek-vl2: Mixture-of-experts vision-language models for advanced multimodal understanding.
\newblock \emph{arXiv preprint arXiv:2412.10302}, 2024.

\bibitem{januspro}
Chen, X., Z.~Wu, X.~Liu, et~al.
\newblock Janus-pro: Unified multimodal understanding and generation with data and model scaling.
\newblock \emph{arXiv preprint arXiv:2501.17811}, 2025.

\bibitem{VLIA}
Lin, J., H.~Yin, W.~Ping, et~al.
\newblock Vila: On pre-training for visual language models.
\newblock In \emph{Proceedings of the IEEE/CVF conference on computer vision and pattern recognition}, pages 26689--26699. 2024.

\bibitem{internvl2-5}
Chen, Z., W.~Wang, Y.~Cao, et~al.
\newblock Expanding performance boundaries of open-source multimodal models with model, data, and test-time scaling.
\newblock \emph{arXiv preprint arXiv:2412.05271}, 2024.

\bibitem{qwen2-5vl}
Bai, S., K.~Chen, X.~Liu, et~al.
\newblock Qwen2. 5-vl technical report.
\newblock \emph{arXiv preprint arXiv:2502.13923}, 2025.

\bibitem{GPT4o}
Hurst, A., A.~Lerer, A.~P. Goucher, et~al.
\newblock Gpt-4o system card.
\newblock \emph{arXiv preprint arXiv:2410.21276}, 2024.

\bibitem{Claude3}
{Anthropic}.
\newblock {The Claude 3 Model Family: Opus, Sonnet, Haiku}.
\newblock Tech. rep., Anthropic, 2024.

\bibitem{Llava-med}
Li, C., S.~J. Wong, S.~Zhang, et~al.
\newblock {LLaVA-Med: Training a Large Language-and-Vision Assistant for Biomedicine in One Day}.
\newblock In \emph{NeurIPS}. 2024.

\bibitem{medflamingo}
Moor, M., Q.~Huang, S.~Wu, et~al.
\newblock Med-flamingo: a multimodal medical few-shot learner.
\newblock In \emph{Machine Learning for Health (ML4H)}, pages 353--367. PMLR, 2023.

\bibitem{QilinMed}
Ye, Q., J.~Liu, D.~Chong, et~al.
\newblock Qilin-med: Multi-stage knowledge injection advanced medical large language model.
\newblock \emph{arXiv preprint arXiv:2310.09089}, 2023.

\bibitem{huatuogptVision}
Chen, J., C.~Gui, R.~Ouyang, et~al.
\newblock Huatuogpt-vision, towards injecting medical visual knowledge into multimodal llms at scale.
\newblock \emph{EMNLP}, 2024.

\bibitem{Meddr}
He, S., Y.~Nie, Z.~Chen, et~al.
\newblock Meddr: Diagnosis-guided bootstrapping for large-scale medical vision-language learning.
\newblock \emph{arXiv e-prints}, pages arXiv--2404, 2024.

\bibitem{VLIA-M3}
Nath, V., W.~Li, D.~Yang, et~al.
\newblock Vila-m3: Enhancing vision-language models with medical expert knowledge.
\newblock \emph{CVPR}, 2025.

\bibitem{Llava-Rad}
Zambrano~Chaves, J.~M., S.-C. Huang, Y.~Xu, et~al.
\newblock A clinically accessible small multimodal radiology model and evaluation metric for chest x-ray findings.
\newblock \emph{Nature Communications}, 16(1):3108, 2025.

\bibitem{MedPLIB}
Huang, X., L.~Shen, J.~Liu, et~al.
\newblock Towards a multimodal large language model with pixel-level insight for biomedicine.
\newblock In \emph{AAAI}, vol.~39, pages 3779--3787. 2025.

\bibitem{biomedclip}
Zhang, S., Y.~Xu, N.~Usuyama, et~al.
\newblock Biomedclip: a multimodal biomedical foundation model pretrained from fifteen million scientific image-text pairs.
\newblock \emph{arXiv preprint arXiv:2303.00915}, 2023.

\bibitem{Surgical-lvlm}
Wang, G., L.~Bai, W.~J. Nah, et~al.
\newblock Surgical-lvlm: Learning to adapt large vision-language model for grounded visual question answering in robotic surgery.
\newblock \emph{ICLR Workshop}, 2025.

\bibitem{GP-VLS}
Schmidgall, S., J.~Cho, C.~Zakka, et~al.
\newblock Gp-vls: A general-purpose vision language model for surgery.
\newblock \emph{arXiv preprint arXiv:2407.19305}, 2024.

\bibitem{Asclepius}
Liu, J., W.~Wang, Y.~Su, et~al.
\newblock A spectrum evaluation benchmark for medical multi-modal large language models.
\newblock \emph{arXiv preprint arXiv:2402.11217}, 2024.

\bibitem{Medifusion}
Sepehri, M.~S., Z.~Fabian, M.~Soltanolkotabi, et~al.
\newblock Mediconfusion: Can you trust your ai radiologist? probing the reliability of multimodal medical foundation models.
\newblock \emph{arXiv preprint arXiv:2409.15477}, 2024.

\bibitem{Medxpertqa}
Zuo, Y., S.~Qu, Y.~Li, et~al.
\newblock Medxpertqa: Benchmarking expert-level medical reasoning and understanding.
\newblock \emph{arXiv preprint arXiv:2501.18362}, 2025.

\bibitem{PMC-VQA}
Zhang, X., C.~Wu, Z.~Zhao, et~al.
\newblock Pmc-vqa: Visual instruction tuning for medical visual question answering.
\newblock \emph{arXiv preprint arXiv:2305.10415}, 2023.

\bibitem{MedTrinity}
Xie, Y., C.~Zhou, L.~Gao, et~al.
\newblock Medtrinity-25m: A large-scale multimodal dataset with multigranular annotations for medicine.
\newblock \emph{arXiv preprint arXiv:2408.02900}, 2024.

\bibitem{Cares}
Xia, P., Z.~Chen, J.~Tian, et~al.
\newblock Cares: A comprehensive benchmark of trustworthiness in medical vision language models.
\newblock \emph{NeurIPS}, 37:140334--140365, 2024.

\bibitem{PSI-AVA}
Valderrama, N., P.~Ruiz~Puentes, I.~Hern{\'a}ndez, et~al.
\newblock Towards holistic surgical scene understanding.
\newblock In \emph{MICCAI}, pages 442--452. Springer, 2022.

\bibitem{Cholec80}
Twinanda, A.~P., S.~Shehata, D.~Mutter, et~al.
\newblock Endonet: a deep architecture for recognition tasks on laparoscopic videos.
\newblock \emph{IEEE transactions on medical imaging}, 36(1):86--97, 2016.

\bibitem{EndoVis18}
Allan, M., S.~Kondo, S.~Bodenstedt, et~al.
\newblock 2018 robotic scene segmentation challenge.
\newblock \emph{arXiv preprint arXiv:2001.11190}, 2020.

\bibitem{Hyperkvasir}
Borgli, H., V.~Thambawita, P.~H. Smedsrud, et~al.
\newblock Hyperkvasir, a comprehensive multi-class image and video dataset for gastrointestinal endoscopy.
\newblock \emph{Scientific data}, 7(1):283, 2020.

\bibitem{Kvasir-Instrucment}
Jha, D., S.~Ali, K.~Emanuelsen, et~al.
\newblock Kvasir-instrument: Diagnostic and therapeutic tool segmentation dataset in gastrointestinal endoscopy.
\newblock In \emph{MultiMedia Modeling: 27th International Conference, MMM 2021, Prague, Czech Republic, June 22--24, 2021, Proceedings, Part II 27}, pages 218--229. Springer, 2021.

\bibitem{Kvasir}
Pogorelov, K., K.~R. Randel, C.~Griwodz, et~al.
\newblock {Kvasir: A multi-class image dataset for computer aided gastrointestinal disease detection}.
\newblock In \emph{ACM MMSys}, pages 166--167. 2017.

\bibitem{Kvasir-capsule}
Smedsrud, P.~H., V.~Thambawita, S.~A. Hicks, et~al.
\newblock {Kvasir-Capsule, a video capsule endoscopy dataset}.
\newblock \emph{Sci. Data}, 8(1):142, 2021.

\bibitem{Gastrovision}
Jha, D., V.~Sharma, N.~Dasu, et~al.
\newblock {GastroVision: A Multi-Class Endoscopy Image Dataset for Computer Aided Gastrointestinal Disease Detection}.
\newblock In \emph{ICML Workshops}. 2023.

\bibitem{KID}
Koulaouzidis, A., D.~K. Iakovidis, D.~E. Yung, et~al.
\newblock {KID project: An internet-based digital video atlas of capsule endoscopy for research purposes}.
\newblock \emph{Endosc. Int. Open}, 5(06):E477--E483, 2017.

\bibitem{WCEBleedGen}
Handa, P., M.~Dhir, A.~Mahbod, et~al.
\newblock {WCEBleedGen: A wireless capsule endoscopy dataset and its benchmarking for automatic bleeding classification, detection, and segmentation}.
\newblock arXiv preprint arXiv:2408.12466, 2024.

\bibitem{SEE-AI}
CapsuleYolo.
\newblock {KyuCapsule Dataset}.
\newblock \url{https://www.kaggle.com/datasets/capsuleyolo/kyucapsule }, n.d.
\newblock Accessed: 2025-04-05.

\bibitem{Kvasir-Seg}
Jha, D., P.~H. Smedsrud, M.~A. Riegler, et~al.
\newblock {Kvasir-SEG: A segmented polyp dataset}.
\newblock In \emph{Multimedia Modeling (MMM)}, pages 30--41. 2020.

\bibitem{CVC-ColonDB}
Bernal, J., J.~S{\'a}nchez, F.~Vilarino.
\newblock Towards automatic polyp detection with a polyp appearance model.
\newblock \emph{Pattern Recognit.}, 45(9):3166--3182, 2012.

\bibitem{ETIS-Larib}
Silva, J., A.~Histace, O.~Romain, et~al.
\newblock Toward embedded detection of polyps in wce images for early diagnosis of colorectal cancer.
\newblock \emph{Comput. Aided Surg.}, 9(2):283--293, 2014.

\bibitem{CVC-ClinicDB}
Bernal, J., F.~J. S{\'a}nchez, G.~Fern{\'a}ndez-Esparrach, et~al.
\newblock Wm-dova maps for accurate polyp highlighting in colonoscopy: Validation vs. saliency maps from physicians.
\newblock \emph{Comput. Med. Imaging Graph.}, 43:99--111, 2015.

\bibitem{CVC-300}
V{\'a}zquez, D., J.~Bernal, F.~J. S{\'a}nchez, et~al.
\newblock A benchmark for endoluminal scene segmentation of colonoscopy images.
\newblock \emph{Journal of healthcare engineering}, 2017(1):4037190, 2017.

\bibitem{EDD2020}
Ali, S., N.~Ghatwary, B.~Braden, et~al.
\newblock Endoscopy disease detection challenge 2020.
\newblock arXiv preprint arXiv:2003.03376, 2020.

\bibitem{SUN-database}
Misawa, M., S.-e. Kudo, Y.~Mori, et~al.
\newblock Development of a computer-aided detection system for colonoscopy and a publicly accessible large colonoscopy video database (with video).
\newblock \emph{Gastrointestinal endoscopy}, 93(4):960--967, 2021.

\bibitem{LDpolypvideo}
Ma, Y., X.~Chen, K.~Cheng, et~al.
\newblock {LDPolypVideo benchmark: A large-scale colonoscopy video dataset of diverse polyps}.
\newblock In \emph{MICCAI}, pages 243--253. 2021.

\bibitem{polypgen}
Ali, S., D.~Jha, N.~Ghatwary, et~al.
\newblock A multi-centre polyp detection and segmentation dataset for generalisability assessment.
\newblock \emph{Scientific Data}, 10(1):75, 2023.

\bibitem{endovis17}
Allan, M., A.~Shvets, T.~Kurmann, et~al.
\newblock 2017 robotic instrument segmentation challenge.
\newblock \emph{arXiv preprint arXiv:1902.06426}, 2019.

\bibitem{dinov2}
Oquab, M., T.~Darcet, T.~Moutakanni, et~al.
\newblock Dinov2: Learning robust visual features without supervision.
\newblock \emph{TMLR}, pages 1--31, 2024.

\bibitem{llavanext}
Liu, H., C.~Li, Y.~Li, et~al.
\newblock Llava-next: Improved reasoning, ocr, and world knowledge, 2024.

\bibitem{qvq}
Team, Q.
\newblock Qvq: To see the world with wisdom, 2024.

\bibitem{hurst2024gpt}
Hurst, A., A.~Lerer, A.~P. Goucher, et~al.
\newblock Gpt-4o system card.
\newblock \emph{arXiv preprint arXiv:2410.21276}, 2024.

\bibitem{grok3}
xAI.
\newblock Grok 3 beta --- the age of reasoning agents, 2025.

\bibitem{radford2021learning}
Radford, A., J.~W. Kim, C.~Hallacy, et~al.
\newblock Learning transferable visual models from natural language supervision.
\newblock In \emph{International conference on machine learning}, pages 8748--8763. PmLR, 2021.

\bibitem{zheng2023judging}
Zheng, L., W.-L. Chiang, Y.~Sheng, et~al.
\newblock Judging llm-as-a-judge with mt-bench and chatbot arena.
\newblock \emph{Advances in Neural Information Processing Systems}, 36:46595--46623, 2023.

\bibitem{lu2022learn}
Lu, P., S.~Mishra, T.~Xia, et~al.
\newblock Learn to explain: Multimodal reasoning via thought chains for science question answering.
\newblock \emph{Advances in Neural Information Processing Systems}, 35:2507--2521, 2022.

\bibitem{grattafiori2024llama}
Grattafiori, A., A.~Dubey, A.~Jauhri, et~al.
\newblock The llama 3 herd of models.
\newblock \emph{arXiv preprint arXiv:2407.21783}, 2024.

\bibitem{chen2024sharegpt4v}
Chen, L., J.~Li, X.~Dong, et~al.
\newblock Sharegpt4v: Improving large multi-modal models with better captions.
\newblock In \emph{European Conference on Computer Vision}, pages 370--387. Springer, 2024.

\bibitem{zhai2023sigmoid}
Zhai, X., B.~Mustafa, A.~Kolesnikov, et~al.
\newblock Sigmoid loss for language image pre-training.
\newblock \emph{arXiv preprint arXiv:2303.15343}, 2023.

\bibitem{yue2024mmmu}
Yue, X., Y.~Ni, K.~Zhang, et~al.
\newblock Mmmu: A massive multi-discipline multimodal understanding and reasoning benchmark for expert agi.
\newblock In \emph{Proceedings of the IEEE/CVF Conference on Computer Vision and Pattern Recognition}, pages 9556--9567. 2024.

\bibitem{liu2024deepseek}
Liu, A., B.~Feng, B.~Xue, et~al.
\newblock Deepseek-v3 technical report.
\newblock \emph{arXiv preprint arXiv:2412.19437}, 2024.

\bibitem{team2023gemini}
Team, G., R.~Anil, S.~Borgeaud, et~al.
\newblock Gemini: a family of highly capable multimodal models.
\newblock \emph{arXiv preprint arXiv:2312.11805}, 2023.

\bibitem{CoT}
Wei, J., X.~Wang, D.~Schuurmans, et~al.
\newblock Chain-of-thought prompting elicits reasoning in large language models.
\newblock In \emph{NeurIPS}, vol.~35, pages 24824--24837. Curran Associates, Inc., 2022.

\bibitem{cot_use01}
Liu, R., Z.~Wu, Y.~Jurals, et~al.
\newblock Mind your step (by step): Chain-of-thought can reduce performance on tasks where thinking makes humans worse, 2025.

\bibitem{cot_use02}
Sprague, Z., Z.~Pu, H.~Peng, et~al.
\newblock To cot or not to cot? chain-of-thought helps mainly on math and symbolic reasoning.
\newblock In \emph{ICLR}. 2025.

\bibitem{cot_use03}
Le, C., Z.~Gong, C.~Wang, et~al.
\newblock Instruction tuning and cot prompting for contextual medical qa with llms.
\newblock \emph{arXiv preprint arXiv:2506.12182}, 2025.

\bibitem{Self-Consistency}
Wang, X., J.~Wei, D.~Schuurmans, et~al.
\newblock Self-consistency improves chain of thought reasoning in language models, 2022.

\bibitem{Self-refine}
Madaan, A., N.~Tandon, P.~Gupta, et~al.
\newblock Self-refine: Iterative refinement with self-feedback.
\newblock In \emph{NeurIPS}, vol.~36. 2023.

\bibitem{medprompt}
Nori, H., Y.~T. Lee, S.~Zhang, et~al.
\newblock Can generalist foundation models outcompete special-purpose tuning? case study in medicine, 2023.

\bibitem{medgraphrag}
Wu, J., J.~Zhu, Y.~Qi.
\newblock Medical graph rag: Towards safe medical large language model via graph retrieval-augmented generation, 2024.

\bibitem{MMed-RAG}
Xia, P., K.~Zhu, H.~Li, et~al.
\newblock Mmed-rag: Versatile multimodal rag system for medical vision language models.
\newblock In \emph{ICLR}. 2025.

\bibitem{lopez2025clinical}
Lopez, I., J.~Min, Z.-Y. Zhao, et~al.
\newblock Clinical entity augmented retrieval for clinical information extraction.
\newblock \emph{npj Digital Medicine}, 8(1):45, 2025.

\end{thebibliography}

\newpage
\appendix
\section*{EndoBench: A Comprehensive Evaluation of Multi-Modal Large Language Models for Endoscopy Analysis}
\addcontentsline{toc}{section}{EndoBench: A Comprehensive Evaluation of Multi-Modal Large Language Models for Endoscopy Analysis} 

\section*{Appendix}
\addcontentsline{toc}{section}{Appendix}

\startcontents[sections]
\printcontents[sections]{l}{1}{\setcounter{tocdepth}{2}}

\newpage
\section{Dataset Details}
\subsection{Involved Datasets}
In this work, we gather 20 public endoscopy datasets from online sources to encompass various endoscopic image types and professional terminology, including Kvasir \cite{Kvasir}, HyperKvasir \cite{Hyperkvasir}, Kvasir-Capsule \cite{Kvasir-capsule}, GastroVision \cite{Gastrovision}, KID \cite{KID}, WCEBleedGen \cite{WCEBleedGen}, SEE-AI \cite{SEE-AI}, Kvasir-Seg \cite{Kvasir-Seg}, CVC-ColonDB \cite{CVC-ColonDB}, ETIS-Larib \cite{ETIS-Larib}, CVC-ClinicDB \cite{CVC-ClinicDB}, CVC-300 \cite{CVC-300}, EDD2020 \cite{EDD2020}, SUN-Database \cite{SUN-database}, LDPolypVideo \cite{LDpolypvideo}, PolypGen \cite{polypgen}, Cholec80 \cite{Cholec80}, EndoVis-17 \cite{endovis17}, EndoVis-18 \cite{EndoVis18}, and PSI-AVA \cite{PSI-AVA}. On the other hand, we further enhance the data diversity by incorporating a private wireless capsule endoscopy image dataset from partner hospitals. All data undergoes the privacy de-identification in accordance with medical ethics requirements.

\begin{table}[hb]
  \centering
  \small
  \caption{Statistics regarding the endoscopic scenarios and dataset information covered by the datasets involved.}
  \label{table:append_datasetinfo}
  \begin{threeparttable}
    \begin{tabular}{cccccc}
      \toprule
      Index & Name & Scenario & Num & Task & Access \\ 
      \midrule
      1 & Kvasir & GS, CS, SE & 8000 & Classification & Open Access \\
      2 & HyperKvasir & GS, CS, SE & 10662 & Classification & Open Access \\
      3 & Kvasir-Capsule & CE & 47238 & Classification & Open Access \\
      4 & GastroVision & GS, CS, SE & 8000 & Classification & Open Access \\
      5 & KID & CE & 2371 & Classification, Segmentation & Restricted Access \\
      6 & WCEBleedGen & CE & 2618 & Classification, Segmentation & Open Access \\
      7 & SEE-AI & GS & 18481 & Classification & Open Access \\
      8 & Kvasir-Seg & CS & 1000 & Segmentation & Open Access \\
      9 & CVC-ColonDB & CS & 380 & Segmentation & Open Access \\
      10 & ETIS-Larib & CS & 196 & Segmentation & Open Access \\
      11 & CVC-ClinicDB & CS & 612 &  Segmentation & Open Access \\
      12 & CVC-300 & CS & 60 & Segmentation & Open Access \\
      13 & EDD2020 & CS & 386 & Classification, Segmentation & Open Access \\
      14 & SUN-Database & CS & 130519 & Classification, Segmentation & Restricted Access \\
      15 & LDPolypVideo & CS & 40266 & Detection & Open Access \\
      16 & PolypGen & CS & 8037 & Segmentation & Open Access \\
      17 & Cholec80 & SE & 8080 & Classification, Segmentation & Open Access \\
      18 & EndoVis-17 & SE & 2235 & Classification, Segmentation & Open Access \\
      19 & EndoVis-18 & SE & 2400 & Classification, Segmentation & Open Access \\
      20 & PSI-AVA & SE & 4471 & Classification, Segmentation & Open Access \\
      21 & WCE2025 & CE & 23447 & Classification & In House \\
      \bottomrule
    \end{tabular}
\begin{flushleft}
  {\footnotesize
    Abbreviation: GS for Gastroscopy, CS for Colonoscopy, CE for Capsule endoscopy, SE for Surgical endoscopy.}
\end{flushleft}
  \end{threeparttable}
\end{table}

WCE2025 is a capsule endoscopy dataset, meticulously curated with all privacy information removed and approved for public use through agreements with relevant hospitals. The dataset, annotated by professional medical experts, comprises a total of 23,447 samples and is designed for three distinct tasks: Landmark Identification with 1,665 samples, Lesion Type Identification with 19,881 samples, and Organ Identification with 1,901 samples. 
\subsection{Task Definition}
To comprehensively evaluate these capacities in MLLMs,~\ourmethod~encompasses 12 clinical tasks with 12 secondary subtasks across 4 major categories for endoscopy analysis. These categories include: (1) \textit{anatomical structure recognition} (organ identification, landmark identification); (2) \textit{lesion analysis and grading} (lesion quantification, lesion type identification, lesion severity grading); (3) \textit{spatial localization and region understanding} (visual grounding, region selection, region recognition); and (4) \textit{surgical workflow and operation analysis} (preoperative assessment, macro phases identification, micro operation analysis, instrument management). 

\subsubsection{Anatomical Structure Recognition}
Anatomical structure recognition is a critical component of endoscopic analysis, enabling the identification of anatomical features within the gastrointestinal tract or other internal structures. This category focuses on two key subtasks:
\begin{itemize}[leftmargin=0.5cm]
    \item \textbf{Landmark Identification (LI).} This task involves detecting and labeling specific anatomical landmarks, such as the pylorus, cardia, or ileocecal valve, which serve as reference points during endoscopic navigation. Accurate landmark identification ensures precise orientation and facilitates diagnostic accuracy.
    \item \textbf{Organ Identification (OI).} This task requires recognizing and classifying entire organs or organ segments visible in endoscopic images, such as the esophagus, stomach, or small/large intestine. Organ identification is essential for contextualizing findings and guiding subsequent clinical decisions.
\end{itemize}

\subsubsection{Lesion Analysis and Grading}
Lesion analysis and grading focus on characterizing abnormalities observed during endoscopy, providing critical information for diagnosis and treatment planning. This category encompasses the following subtasks:
\begin{itemize}[leftmargin=0.5cm]
    \item \textbf{Lesion Quantification (LQ).}: This task mainly involves measuring the number of polyps, which is crucial for assessing the extent of disease, monitoring progression, and guiding treatment decisions. Accurate quantification provides essential data for evaluating health risks and planning effective interventions.
    \item \textbf{Lesion Type Identification (LT)}: This task includes:
    \begin{itemize}[leftmargin=0.5cm]
        \item \textit{Lesion Classification (LC).}: Categorizing lesions into broad types, such as erosions, ulcers, esophagitis, or angiectasia, based on their visual characteristics.
        \item \textit{Polyp Type Classification (PT).} Specifically identifying polyp types, such as adenomatous, hyperplastic, or serrated, which have distinct clinical implications.
    \end{itemize}
    \item \textbf{Lesion Severity Grading (LS).} This task involves assessing the severity of ulcerative colitis through colonoscopy, using the UCEIS Mayo Score to classify mucosal inflammation into four levels (0: inactive, 1: mild, 2: moderate, 3: severe) and additional in-between classes (0-1, 1-2, 2-3) to account for observer variation and nuanced disease presentations. This task is critical for determining disease extent, guiding treatment, and monitoring progression.
\end{itemize}

\subsubsection{Surgical Workflow and Operation Analysis}
Surgical workflow and operation analysis focus on understanding and optimizing endoscopic procedures, from preoperative planning to intraoperative management. This category includes the following tasks:

\begin{itemize}[leftmargin=0.5cm]
    \item \textbf{Preoperative Assessment (PA).} This task focuses on evaluating bowel cleanliness prior to surgical or endoscopic procedures, primarily through the Boston Bowel Preparation Scale (BBPS). This task involves scoring the quality of mucosal views in endoscopic images or videos, using only the BBPS 0-1 (poorly prepared, with significant stool or fluid obstructing the view) and BBPS 2-3 (well-prepared, with minimal or no residue, including perfectly clean BBPS 3 cases) classes, with the impacted stool class excluded.
    \item \textbf{Macro Phases Identification (MP)}: This task includes:
    \begin{itemize}[leftmargin=0.5cm]
        \item \textit{Surgical Phase Recognition (SP).} Identifying distinct phases of an endoscopic procedure, such as clipping and cutting, gallbladder dissection, preparation, gallbladder retraction, and others. This task is essential for enhancing procedural efficiency, supporting surgical training, and enabling automated analysis of surgical workflows.
        \item \textit{Surgical Step Recognition (SS).} Identifying specific steps within an endoscopic procedure phase, such as inserting the prostate into a retrieval bag, prostate dissection until the levator ani, or vascular pedicle control. This task is essential for tracking procedural progress, enhancing surgical precision, and supporting training and automated surgical workflow analysis.
    \end{itemize}
    \item \textbf{Micro Operation Analysis (MO).} This task includes:
    \begin{itemize}[leftmargin=0.5cm]
        \item \textit{Surgical Action Recognition (SA).} Detecting specific actions performed by a particular instrument during endoscopic interventions, such as suction, tissue manipulation, suturing, or idle states. This task is crucial for analyzing instrument-specific activities, optimizing procedural efficiency, and supporting surgical training and automation.
        \item \textit{Treatment Intervention Recognition (TI).} This task involves identifying and classifying therapeutic interventions performed during endoscopic procedures to address detected anomalies, such as lesion or polyp removal. It includes recognizing actions like polyp resection, biopsy of resection margins or sites, and the use of indigo carmine injection to enhance lesion demarcation, where the blue color beneath the dyed, lifted polyp highlights accurate polyp margins. This task is critical for ensuring precise treatment execution and supporting procedural documentation and analysis.
    \end{itemize}
    \item \textbf{Instrument Management (IM).} This task includes:
    \begin{itemize}[leftmargin=0.5cm]
        \item \textit{Instrument Counting (IC).} This subtask involves identifying and counting the number of distinct types of instruments used during an endoscopic procedure. It requires recognizing various instruments present in the endoscopic images or videos to provide an accurate count of different tools, such as forceps, scissors, or suction devices.
        \item \textit{Instrument Presence Verification (IP).} This subtask involves determining whether a specific instrument, such as forceps, scissors, or a suction device, is present in endoscopic images or videos. It requires analyzing the visual data to confirm the presence or absence of the specified instrument, which is crucial for ensuring appropriate tool usage, tracking procedural steps, and supporting surgical workflow management.
    \end{itemize}
\end{itemize}

\subsubsection{Spatial Localization and Region Understanding}
Spatial localization and region understanding enable precise mapping and interpretation of regions of interest within endoscopic images. This category includes the following subtasks:
\begin{itemize}[leftmargin=0.5cm]
    \item \textbf{Visual Grounding (VG).} This task involves associating textual descriptions or clinical queries with specific regions in endoscopic images to accurately identify relevant features, such as anatomical structures, lesions, or abnormalities. The task requires selecting the correct coordinate location from four candidate options to ensure precise localization of these features in the images.
    \item \textbf{Region Selection (RS)}: This task involves identifying and selecting key regions of interest, such as abnormal tissue areas or anatomical structures, in endoscopic images for further analysis or intervention. The task requires choosing the correct region from four candidate regions, each marked with a different color, to ensure accurate localization of the critical area.
    \item \textbf{Region Recognition (RR).} This task involves identifying and classifying specific structures or lesions in endoscopic images using various visual prompt methods to determine what a given region represents. The task includes the following subtasks:
    \begin{itemize}[leftmargin=0.5cm]
        \item \textit{Bounding Box Region Recognition.} A rectangular bounding box is overlaid on the endoscopic image to highlight a specific area, and the task is to identify the structure or lesion within that boxed region.
        \item \textit{Contour-based Region Recognition.} Precise boundary outlines are drawn on the image to determine the corresponding structure or lesion.
        \item \textit{Multi-region Recognition.} Multiple regions within a single endoscopic frame are marked with distinct colors, and the task is to identify the structure or lesion associated with a specified color.
        \item \textit{Coordinate-based Region Recognition.} Specific coordinates (e.g., $[x1, y1, x2, y2]$) are provided in the question, and the task is to identify the structure or lesion at that precise location, facilitating integration with coordinate-based systems.
    \end{itemize}
\end{itemize}

\subsubsection{Summarization of Clinically-Grounded Tasks}
The 12 tasks in EndoBench are systematically designed to mirror the end-to-end clinical workflow of an endoscopic procedure, which encompasses three critical phases: diagnostic assessment, surgical planning, and therapeutic implementation. Each task is mapped to a specific clinical need within this progression.

The workflow begins with the diagnostic assessment phase. When a patient undergoes an endoscopic evaluation, the procedure involves a systematic examination of the gastrointestinal tract. Organ Identification provides real-time anatomical orientation as the endoscope navigates between the esophagus, stomach, and duodenum. Concurrently, Landmark Identification recognizes critical structures, such as the pylorus or ampulla of Vater, to ensure a complete and thorough examination. Upon detecting an abnormality, Lesion Type Identification characterizes its pathology (e.g., polyps, ulcers, inflammation), Lesion Quantification counts multiple instances for treatment planning, and Lesion Severity Grading applies standardized scoring systems (e.g., Mayo Score, BBPS) to guide therapeutic decisions.

Following diagnosis, the process transitions to the surgical planning phase, where detailed preparation for intervention is critical. In this stage, Visual Grounding provides precise coordinates for surgical navigation systems, Region Selection facilitates accurate target localization, Region Recognition characterizes specific anatomical structures within an area of interest, and Preoperative Assessment evaluates key safety parameters before an intervention begins.

The final stage is therapeutic implementation, where the focus shifts to real-time procedural execution and quality assurance. Macro Phases Identification offers high-level workflow guidance by recognizing distinct surgical stages (e.g., dissection, resection, suturing). To enhance patient safety, Instrument Management monitors the presence of tools to prevent retained equipment. Finally, Micro Operation Analysis provides a granular assessment of surgical skill and quality by analyzing specific instrument functions, such as grasping, cutting, and cauterization.

By structuring the tasks around this comprehensive workflow, EndoBench ensures that MLLMs are evaluated on capabilities that directly translate to improved patient care, enhanced surgical navigation, and more informed clinical decision-making in real-world endoscopic practice. This end-to-end mapping underscores the clinical value and practical relevance of the benchmark.
\subsection{Construction Process of QA Pairs}
\textbf{Task Templates.} To ensure a comprehensive and varied evaluation of model performance, we develop 5 to 8 distinct question templates for each task. These templates are designed to cover a range of scenarios and complexities, enabling robust testing of the model's capabilities. The table below outlines the specific prompts associated with each task, providing a structured framework for generating diverse and targeted questions.
\begin{tcolorbox}[
  colframe=gray!50, 
  colback=gray!10, 
  title=Prompt for Answer Evaluation, 
  fonttitle=\bfseries, 
  boxrule=0.5mm, 
  arc=4mm, 
  left=3mm, 
  right=3mm, 
  top=3mm, 
  bottom=3mm, 
  fontupper=\small\ttfamily, 
  breakable
]
\textbf{Landmark Identification:} \\
What anatomical landmark is highlighted in this image? \\
Which anatomical landmark is visible in this image? \\
Can you identify the anatomical landmark in this image? \\
Identify the anatomical landmark in this image. \\
What is the name of the anatomical landmark in this image? \\
Which anatomical structure is shown in this image? \\
Can you identify the anatomical structure in this image? \\
Identify the anatomical structure in this image. \\
What is the name of the anatomical feature marked in this image? \\

\textbf{Organ Identification:} \\
What organ is shown in this image? \\
Which part of the digestive system is depicted in this image? \\
Can you identify the organ in this image? \\
This image shows characteristic features of which digestive organ? \\
Can you identify the digestive organ in this image? \\
Identify the digestive organ in this image. \\
What is the name of the digestive organ shown in this image? \\

\textbf{Lesion Quantification:} \\
Based on the image provided, how many polyps are present in the image? \\
Can you identify the number of polyps in this image? \\
Please identify the number of polyps shown in the image. \\
Given the endoscopic image, can you determine the number of polyps? \\
Identify the number of polyps in this endoscopic image. \\

\textbf{Lesion Classification:} \\
Is there any abnormality visible in this image? If so, describe the type of abnormality. \\
Based on this endoscopic image, what type of abnormal finding can be identified? \\
Does this endoscopic image show any abnormalities? If yes, please specify the type. \\
Are there any abnormal findings in this image? If present, what type of abnormality is it? \\
Please examine this image and indicate if there are any abnormalities. If so, what kind? \\
Review this image and state if there are any abnormalities. If found, specify the type. \\
Check this image for any abnormalities. If detected, what type of abnormality is present? \\
Analyze this image and report if there are any abnormalities. If yes, describe the type. \\
Evaluate this image for abnormalities. If any are found, what type are they? \\

\textbf{Lesion Severity Grading:} \\
What is the Mayo Score for ulcerative colitis in this endoscopic image? \\
Can you determine the Mayo Score for ulcerative colitis in this endoscopic image? \\
Based on the Mayo Score, what score would you give this endoscopic image for ulcerative colitis? \\
According to the Mayo Score, how would you score this endoscopic image for ulcerative colitis? \\
What score does this endoscopic image achieve when assessed using the Mayo Score for ulcerative colitis? \\

\textbf{Polyp Type Classification:} \\
Based on the image provided, identify the histopathological type of the lesion. \\
Can you identify the histopathological type of the lesion in this image? \\
Please identify the histopathological type of the lesion shown in the image. \\
Given the endoscopic image, can you determine the colorectal lesion type? \\
What type of colorectal lesion is depicted in the image? \\

\textbf{Visual Grounding:}\\
Could you give the location of the \{lesion\_type\} in this endoscopic image? \\
Please specify the coordinates of the \{lesion\_type\} in this endoscopic image. \\
Could you specify the location of the \{lesion\_type\} in this endoscopic image? \\
Please give the coordinates of the \{lesion\_type\} in this endoscopic image. \\
Please specify the location of the \{lesion\_type\} in the image. \\
Could you identify the coordinates of the \{lesion\_type\} in the image? \\

\textbf{Region Selection:}\\
In the given image, which color box best represents the area of the \{lesion\_type\}? \\
In the provided image, which color box best indicates the location of the \{lesion\_type\}? \\
In this endoscopic image, which color box best highlights the \{lesion\_type\}? \\
Which color box in the image best describes the \{lesion\_type\}? \\
Which color box in this endoscopic image best represents the \{lesion\_type\}? \\

\textbf{Bounding Box Region Recognition:}\\
Which option best describes the region marked by the rectangle in the endoscopy image? \\
In the endoscopy image, which option best describes the region marked by the rectangle? \\
Which option best describes the region highlighted by the rectangle in the endoscopy image? \\
In this endoscopy image, which option best describes the highlighted region marked by the rectangle?\\
In this endoscopy image, which surgical instrument is indicated by the \{color\} bounding box? \\
Given the endoscopy image, what surgical instrument is shown in the \{color\} bounding box? \\
Based on the endoscopy image, identify the surgical instrument in the \{color\} bounding box. \\
Which surgical instrument corresponds to the \{color\} bounding box in this endoscopy image? \\
Determine the surgical instrument in the \{color\} bounding box from the endoscopy image. \\

\textbf{Contour-based Region Recognition:}\\
Which option best describes the region marked by the contour in the endoscopy image? \\
In the endoscopy image, which option best describes the region marked by the contour? \\
Which option best describes the region highlighted by the contour in the endoscopy image? \\
In this endoscopy image, which option best describes the highlighted region marked by the contour? \\

\textbf{Multi-region Recognition:}\\
Which option best describes the region marked by the \{color\} rectangle in the endoscopy image? \\
In the endoscopy image, which option best describes the region marked by the \{color\} bounding box? \\
Which option best describes the region highlighted by the \{color\} rectangle in the endoscopy image? \\
In this endoscopy image, which option best describes the highlighted region marked by the \{color\} bounding box? \\
Which option best describes the region marked by the \{color\} contour in the endoscopy image? \\
In the endoscopy image, which option best describes the region marked by the \{color\} contour? \\
Which option best describes the region highlighted by the \{color\} contour in the endoscopy image? \\
In this endoscopy image, which option best describes the highlighted region marked by the \{color\} contour? \\

\textbf{Coordinate-based Region Recognition:}\\
Could you identify the findings in location at [\{x1\}, \{y1\}, \{x2\}, \{y2\}] in this endoscopic image? \\
What type of finding can be identified at [\{x1\}, \{y1\}, \{x2\}, \{y2\}] in this endoscopic image? \\
Please specify the findings at [\{x1\}, \{y1\}, \{x2\}, \{y2\}] in this endoscopic image. \\
Could you specify the findings at [\{x1\}, \{y1\}, \{x2\}, \{y2\}] in this endoscopic image? \\
Analyze this image and specify the findings at [\{x1\}, \{y1\}, \{x2\}, \{y2\}]. \\
What surgical instrument is located within the bounding box at coordinates [\{x1\}, \{y1\}, \{x2\}, \{y2\}] in this endoscopy image? \\
Identify the surgical instrument inside the rectangle at coordinates [\{x1\}, \{y1\}, \{x2\}, \{y2\}] in this endoscopy image. \\
Which surgical instrument is within the coordinates [\{x1\}, \{y1\}, \{x2\}, \{y2\}] in this endoscopy image? \\
In this endoscopy image, what is the surgical instrument at the bounding box [\{x1\}, \{y1\}, \{x2\}, \{y2\}]? \\
Determine the surgical instrument located at coordinates [\{x1\}, \{y1\}, \{x2\}, \{y2\}] in this endoscopy image. \\

\textbf{Preoperative Assessment:}\\
What is the Boston Bowel Preparation Scale (BBPS) score for this endoscopic image? \\
Can you determine the score for this endoscopic image based on the Boston Bowel Preparation Scale? \\
Based on the Boston Bowel Preparation Scale, what score would you give this endoscopic image? \\
According to the Boston Bowel Preparation Scale, how would you score this endoscopic image? \\
What score does this endoscopic image achieve when assessed using the Boston Bowel Preparation Scale? \\

\textbf{Surgical Phase Recognition:}\\
This is an endoscopy image. Which surgical phase is currently being performed? \\
Given this endoscopy image, can you identify the ongoing surgical phase? \\
Based on the endoscopy image provided, what surgical phase is depicted? \\
Looking at this endoscopy image, which surgical phase does it correspond to? \\
In the context of this endoscopy image, determine the current surgical phase. \\

\textbf{Surgical Step Recognition:}\\
Given this endoscopy image. Which surgical step is being performed? \\
In the endoscopy image, what surgical step is currently underway? \\
Based on this endoscopy image, can you identify the surgical step? \\
Which surgical step does this endoscopy image correspond to? \\
From the endoscopy image provided, determine the surgical step. \\

\textbf{Surgical Action Recognition:}\\
In this endoscopy image, what is the state of the \{instrument\}? \\
What surgical action is the \{instrument\} performing in this endoscopy image? \\
Identify the state of the \{instrument\} in this endoscopy image. \\
What is the \{instrument\} doing in this endoscopy image? \\
Determine the action of the \{instrument\} in this endoscopy image. \\

\textbf{Treatment Intervention Recognition:}\\
What is the therapeutic intervention in this endoscopic image? \\
Can you identify the therapeutic intervention in this image? \\
Identify the therapeutic intervention in this image. \\
Which therapeutic intervention is shown in this image? \\
Which therapeutic intervention is performed in this image? \\

\textbf{Instrument Counting:}\\
This is an endoscopy image. How many distinct types of surgical instruments can be identified? \\
In the endoscopy image provided, which option correctly states the number of unique surgical instrument categories? \\
Based on this endoscopy image, how many different classifications of surgical instruments are visible? \\
From the endoscopy image, can you determine the number of unique surgical instrument varieties present? \\
Which option accurately reflects the count of distinct surgical instrument types in this endoscopy image? \\

\textbf{Instrument Presence Verification:}\\
In this endoscopy image, is \{instrument\} present during \{phase\}? \\
Based on this endoscopy image, is \{instrument\} used in \{phase\}? \\
Does this endoscopy image show \{instrument\} during \{phase\}? \\
In the context of this endoscopy image, is \{instrument\} visible in \{phase\}? \\
Is \{instrument\} present in this endoscopy image during \{phase\}? \\
\end{tcolorbox}

\textbf{Refined QA.} To improve dataset variety and thoroughly assess the capabilities of Multimodal Large Language Models (MLLMs), we employ the GPT-4o-mini API to rephrase questions from the original QA pairs. The following is the prompt provided to GPT-4o-mini:
\begin{tcolorbox}[
  colframe=gray!50, 
  colback=gray!10, 
  title=Prompt for Refined QA, 
  fonttitle=\bfseries, 
  boxrule=0.5mm, 
  arc=4mm, 
  left=3mm, 
  right=3mm, 
  top=3mm, 
  bottom=3mm, 
  fontupper=\small\ttfamily, 
  breakable, 
  parbox=false, 
  width=\textwidth
]
Please rewrite the following question text to make its expression more diverse, while keeping the core meaning unchanged. The question pertains to an endoscopic image,
so please incorporate knowledge from the medical field. Use varied sentence structures and appropriate synonyms, avoiding direct repetition of the original sentence. \\
If possible, include professional expressions commonly used in the medical domain. \\
If you need a reference example, here is a sample question and its rewritten version: \\
Original question:\\
`What is the lesion in this endoscopic image?' \\
Rewritten question: \\
`What type of abnormality might the region observed in this endoscopic image represent?' \\
Now, please rewrite the following question text: \\
The original question text is as follows: \\
\{report\} \\
and the answer options are as follows: \\
\{options\} \\
Please only give one rewritten version of the question like <question> the rewritten question</question>, and DO NOT add any other content like options or answers.
\end{tcolorbox}

\subsection{Well-categorized Data Structure}

In this work, we construct~\ourinstruct~dataset, yielding 446,535 VQA pairs, which is the current largest endoscopic instruction-tuning collection. Fig. \ref{fig:sup_distribution} shows the distribution of the~\ourinstruct~dataset. The dataset exhibits an imbalanced class distribution, with normal samples dominating the Lesion Type Identification task. To address this issue and ensure a more robust evaluation, a smaller, curated subset of the dataset is selected. From~\ourinstruct, we extract representative pairs that undergo rigorous clinical review, resulting in our final~\ourmethod~dataset of 6,832 clinically validated VQA pairs. The construction pipeline is shown in Section 3.1.
\textbf{\begin{figure}[t]
  \centering
   \includegraphics[width=\linewidth]{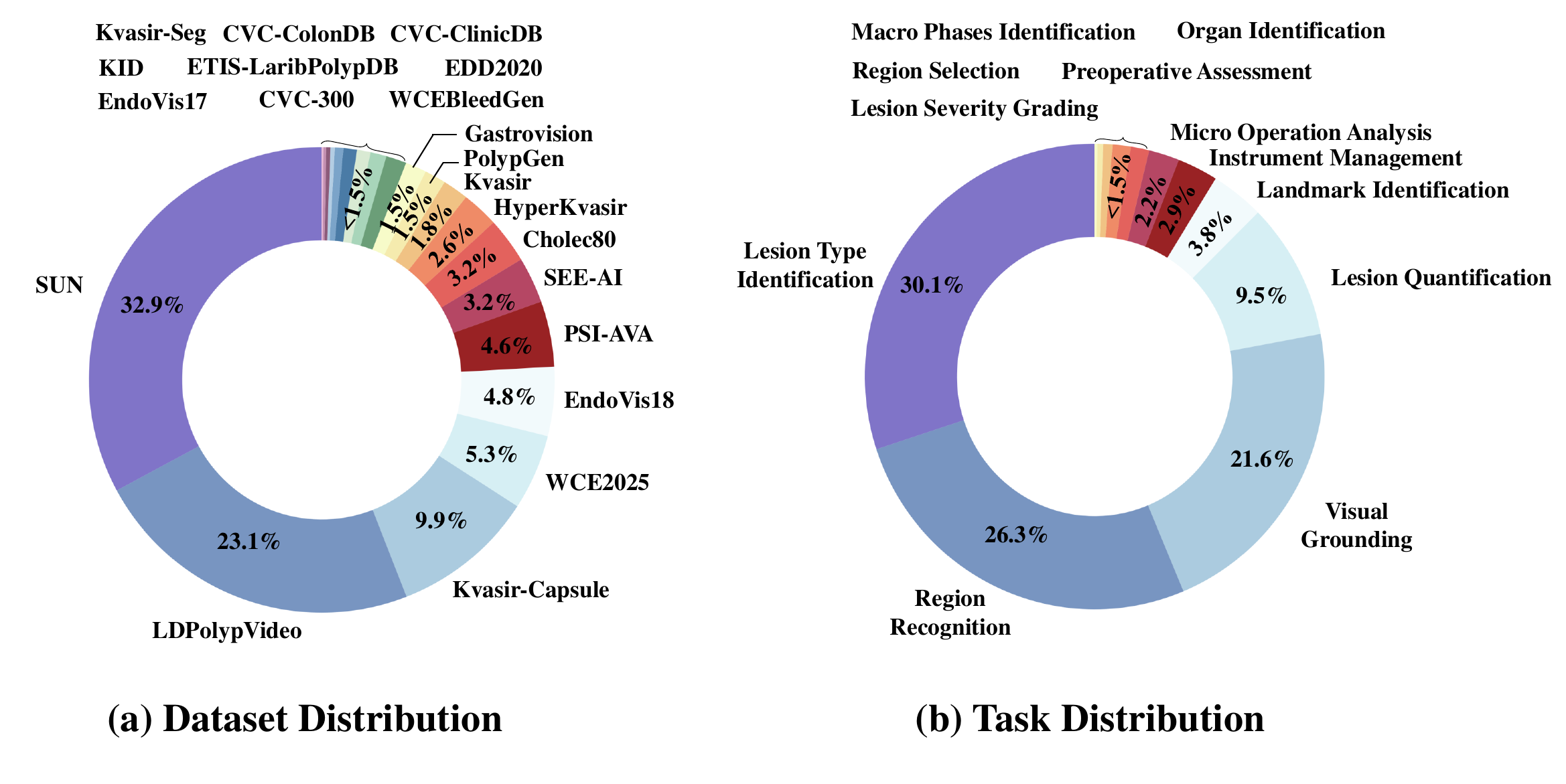}
   \caption{Data distribution of the  ~\ourinstruct~dataset.}
   \label{fig:sup_distribution}
\end{figure}    
}

Regarding surgical procedures, the surgical endoscopy subset includes four representative surgery types:
\begin{itemize}
    \item Abdominal porcine surgeries (30.77\%) from EndoVis17 and EndoVis18,
    \item Human laparoscopic cholecystectomy (16.05\%) from Cholec80,
    \item Human radical prostatectomy (41.98\%) from PSI-AVA, and
    \item Endoscopic mucosal resection (11.20\%) from Kvasir, HyperKvasir, and GastroVision.
\end{itemize}
\section{Evaluation}

\subsection{Evaluated Models}
This study evaluates 23 distinct Multi-modal Large Language Models (MLLMs), spanning open-source architectures, domain-specific medical models, and closed-source general-purpose models accessed via proprietary APIs. 

\begin{itemize}[leftmargin=0.5cm]
    \item \textbf{LLaVA~\cite{Llava}}: This series employs an end-to-end training framework, integrating CLIP-based~\cite{radford2021learning} vision encoders with large language models for robust visual and linguistic understanding. LLaVA-v1.5-7B~\cite{Llava} and LLaVA-v1.5-13B~\cite{Llava} use a Vicuna~\cite{zheng2023judging} backbone, achieving strong performance on benchmarks like Science QA~\cite{lu2022learn}. LLaVA-llama3-8B~\cite{Llava} and LLaVA-Next-Llama3-8B~\cite{llavanext}, built on the 8-billion-parameter LLaMA-3 model~\cite{grattafiori2024llama}, enhance visual reasoning and OCR capabilities for diverse multimodal applications.
    \item \textbf{CogVLM-Chat-7B~\cite{CogVLM}}: Combining a vision transformer encoder, an MLP adapter, and a visual expert module, this model enables deep fusion of visual and linguistic features, excelling in tasks like image captioning and visual question answering with a pretrained language model backbone.
    \item \textbf{ShareGPT-4v-7B~\cite{sharegpt4v}}: This open-source chatbot is trained by fine-tuning a CLIP vision tower and LLaMA/Vicuna on GPT4-Vision-assisted ShareGPT4V data~\cite{chen2024sharegpt4v} and LLaVA instruction-tuning data. Leveraging the ShareGPT4V dataset, it captures detailed visual information, including world knowledge and spatial relationships, enhancing performance across multimodal benchmarks through supervised fine-tuning.
    \item \textbf{Qwen2.5VL~\cite{qwen2-5vl}}: With 3B, 7B, and 72B parameter variants, this series introduces dynamic resolution and frame-rate sampling for video understanding. It excels in object localization, structured output generation for documents like invoices, and advanced visual recognition, supported by an optimized vision encoder with window attention.
    \item \textbf{Janus-Pro-7B~\cite{januspro}}: This model employs a novel autoregressive framework that unifies multimodal understanding and generation. It addresses the limitations of previous approaches by decoupling visual encoding into separate pathways while utilizing a single, unified transformer architecture. Using a SigLIP-L encoder~\cite{zhai2023sigmoid} for 384x384 image inputs and a specialized tokenizer for generation, it supports efficient high-resolution image processing.
    \item \textbf{InternVL2.5~\cite{internvl2-5}}: Comprising 8B and 38B parameter variants, this series enhances multimodal capabilities with optimized training strategies and high-quality data, supporting complex visual-language interactions and high-resolution image processing.
    \item \textbf{QvQ-72B~\cite{qvq}}: Focused on visual reasoning, this experimental model achieves a 70.3\% score on the MMMU benchmark~\cite{yue2024mmmu}, excelling in multidisciplinary understanding, mathematical reasoning, and Olympiad tasks, supporting single-round dialogues and image outputs.
    \item \textbf{MedDr-80B~\cite{Meddr}}: This model handles diverse medical imaging modalities, including radiology, pathology, and endoscopy, using a diagnosis-guided bootstrapping strategy to create high-quality datasets, boosting performance in visual question answering and medical report generation.
    \item \textbf{LLaVA-Med-7B~\cite{Llava-med}}: A biomedical adaptation of LLaVA, this model is fine-tuned with a curriculum learning approach, aligning biomedical vocabulary with figure-caption pairs for efficient handling of medical image queries and multimodal conversations.
    \item \textbf{HuatuoGPT-Vision~\cite{huatuogptVision}}: Available in 7B and 34B variants, this series leverages the PubMedVision dataset~\cite{huatuogptVision} with 1.3 million medical VQA samples, significantly improving medical multimodal benchmarks and enhancing clinical visual understanding.
    \item \textbf{ColonGPT~\cite{colongpt}}: Specialized for colonoscopic image analysis, this model integrates a SigLIP vision encoder and a compact language model, enabling precise diagnostics in gastroenterology through multimodal processing.
    \item \textbf{DeepSeek-V3~\cite{liu2024deepseek}}: Built on a Mixture-of-Experts architecture with 671B parameters, this proprietary model delivers efficient inference for high-resolution image processing and complex reasoning across multimodal tasks.
    \item \textbf{Grok-3~\cite{grok3}}: Designed for versatile multimodal interactions, this model supports text and image inputs with robust reasoning capabilities, optimized for real-time performance across multiple platforms.
    \item \textbf{Claude-3.7-Sonnet~\cite{Claude3}}: A hybrid reasoning model, it excels in content generation, data analysis, and visual reasoning, offering advanced capabilities for diverse multimodal and clinical applications.
    \item \textbf{GPT-4o~\cite{hurst2024gpt}}: A flagship multimodal model, it seamlessly integrates text and image processing, supporting a wide range of tasks requiring robust visual and linguistic understanding.
    \item \textbf{Gemini-2.5-Pro~\cite{team2023gemini}}: Based on a sparse Mixture-of-Experts Transformer, this model enhances complex reasoning, processing inputs from text, images, and code repositories with improved training stability and performance.
\end{itemize}

\subsection{Detailed Setup}
All model weights are obtained from their official repositories on Hugging Face to ensure consistency and reliability. The evaluation is carried out under a zero-shot learning paradigm, where no task-specific training data or in-context examples are provided to the models. 

We adopt a closed-set evaluation protocol to facilitate systematic performance assessment. In this setup, each task is presented as a multiple-choice question, where models are required to select the correct answer from a predefined set of options. The adoption of a closed-set evaluation is particularly well-suited to the medical domain, where diagnostic reasoning often involves distinguishing between highly similar conditions or subtly differentiated concepts. Open-ended generation in such contexts may lead to semantically plausible yet technically incorrect responses, making it difficult to assess precise comprehension. Predictions are evaluated using exact matching against ground truth labels, and we report the accuracy at the task-wise and overall levels. The following is the prompt for evaluation: 
\begin{tcolorbox}[
  colframe=gray!50, 
  colback=gray!10, 
  title=Prompt for Evaluation, 
  fonttitle=\bfseries, 
  boxrule=0.5mm, 
  arc=4mm, 
  left=3mm, 
  right=3mm, 
  top=3mm, 
  bottom=3mm, 
  fontupper=\small\ttfamily, 
  breakable, 
  parbox=false, 
  width=\textwidth
]
USER: <image>\\
\{Question\}: \\
A. \{optionA\} \\
B. \{optionB\} \\
\ldots \\
Please select the correct answer from the options above. \\
ASSISTANT:
\end{tcolorbox}

During the evaluation, we observe that certain MLLMs, especially those tailored for medical applications, struggle with the following instructions, often failing to generate responses in the expected format (e.g., selecting an option letter). This does not necessarily reflect a lack of domain knowledge, but rather a limitation in processing structured question formats. To address this, we employ Qwen2.5VL-72B \cite{qwen2-5vl} as an auxiliary judger to extract the most plausible answer from the model's response when the intended choice is ambiguous or missing. If no valid answer can be identified, the sample is treated as an error. The following is the prompt for evaluation: 

\begin{tcolorbox}[
  colframe=gray!50, 
  colback=gray!10, 
  title=Prompt for LLM Judger, 
  fonttitle=\bfseries, 
  boxrule=0.5mm, 
  arc=4mm, 
  left=3mm, 
  right=3mm, 
  top=3mm, 
  bottom=3mm, 
  fontupper=\small\ttfamily, 
  breakable, 
  parbox=false, 
  width=\textwidth
]
You are tasked with evaluating the correctness of a model's output by comparing it to the ground-truth answer. Extract a plausible answer from the model's output. If no valid answer can be extracted, mark the output as incorrect. Compare the extracted answer (or the original output if extraction is unnecessary) with the ground-truth answer for an exact match. \\
Output: Respond with "Yes" if the model's answer exactly matches the ground-truth answer, or "No" if it does not match or no valid answer could be extracted. \\
Model Output: \{model\_output\} \\
Ground-Truth Answer: \{ground\_truth\_answer\} \\
Provide your response as either "Yes" or "No".
\end{tcolorbox}

\textbf{Human Study.} To establish a benchmark for performance, the study includes an evaluation of human clinicians. We randomly select 255 questions from our~\ourmethod~across all the sub-tasks except coordinate-related tasks, due to the precise coordinate format being unsuitable for intuitive human judgment. Each sub-task may include 5, 10, or 15 samples. Two certified clinicians with expertise in endoscopy independently finished the selected questions. Their scores are averaged for each task to provide a reference standard for comparison. 



\subsection{Additional Results}
In this section, we will provide the complete quantitative results of our experiments. 

Table \ref{tab:subtask_comparison} presents the performance of various MLLMs across 12 subtasks within the \ourmethod framework. Notably, in the Polyp Type Classification task, ColonGPT \cite{colongpt} significantly outperforms other models, achieving an impressive accuracy of 57.60\%. In contrast, competing models exhibit suboptimal performance, primarily due to their limited domain-specific knowledge in this area.

\begin{figure}[t]
  \centering
   \includegraphics[width=\linewidth]{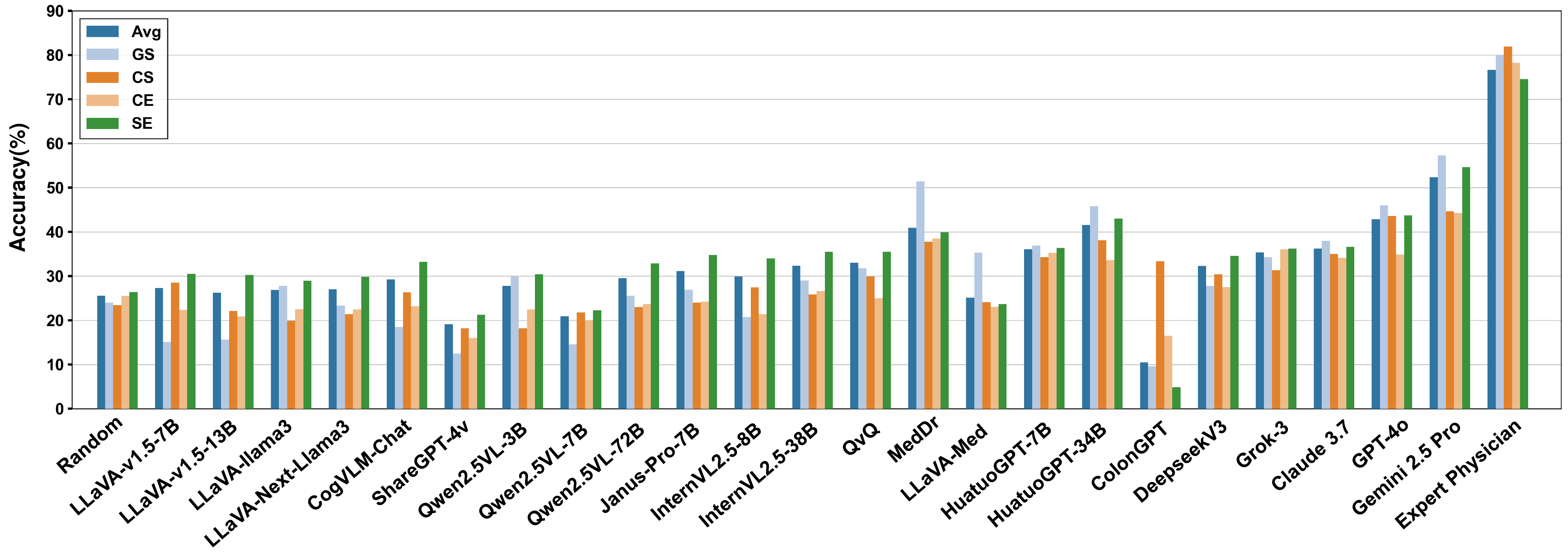}
   \caption{Performance comparison across 4 endoscopic scenarios in~\ourmethod~among existing MLLMs.}
   \label{fig:sup_fig1}
\end{figure}

\begin{figure}[t]
  \centering
   \includegraphics[width=\linewidth]{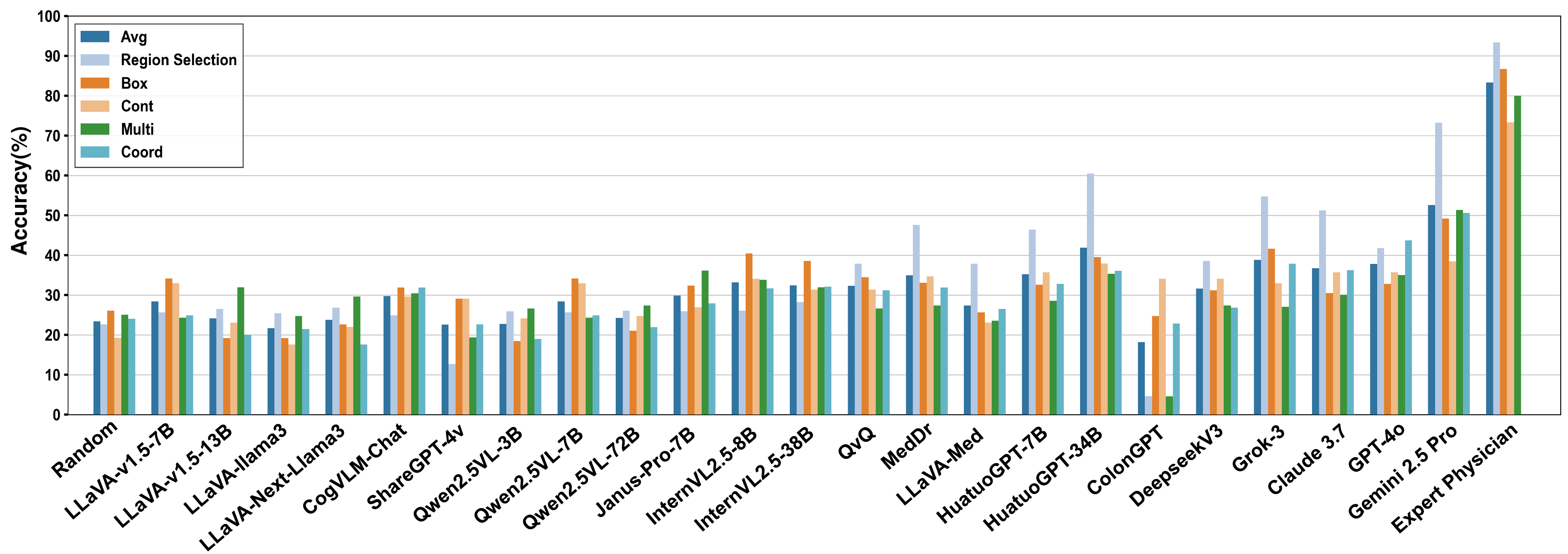}
   \caption{Performance comparison across 5 different visual prompts in~\ourmethod~among existing MLLMs.}
   \label{fig:sup_fig2}
\end{figure}

Fig. \ref{fig:sup_fig1} and \ref{fig:sup_fig2} illustrate the performance comparisons of existing MLLMs across 4 endoscopic scenarios and 5 levels of visual prompting granularity within the~\ourmethod, respectively. The results reveal that open-source general-purpose models tend to perform relatively better on surgical images compared to medical-domain-specific models and proprietary models, where the performance advantage is less pronounced. However, across all endoscopic scenarios and visual prompting granularities, a consistent performance gap exists between all evaluated models and expert physicians. This observation underscores the challenges in achieving expert-level proficiency in medical image analysis and highlights the need for further advancements in model design and training to bridge this gap.

\begin{table}[]
\centering
\caption{Results of different MLLMs on 12 subtasks in~\ourmethod. The best-performing model in each category is \textbf{in-bold}, and the second best is \underline{underlined}.}
\label{tab:subtask_comparison}
\setlength{\tabcolsep}{1.5pt}
\begin{adjustbox}{width=0.95\textwidth,center}
\begin{tabular}{lcccccccccccc}
\toprule
MLLMs & LC & PT & SP & SS & SA & TI & IC & IP & Box & Cont & Mul & Coor \\
\midrule
Random & 24.50 & 19.20 & 25.71 & 25.52 & 26.60 & 29.18 & 25.63 & 48.98 & 26.10 & 19.23 & 25.10 & 24.02 \\
Physician & 86.67 & 70.00 & 53.33 & 66.67 & 85.00 & 66.67 & 90.00 & 73.33 & 86.67 & 73.33 & 80.00 & - \\
\rowcolor{orange!20}
\multicolumn{13}{c}{\textbf{Open-Source MLLMs}} \\ 
Llava-v1.5-7B & 21.29 & 22.80 & 24.86 & 23.71 & 54.29 & 23.61 & 30.15 & 46.94 & 34.18 & 32.97 & 24.33 & 24.94 \\
Llava-v1.5-13B & 19.68 & 20.00 & 19.43 & 25.00 & 55.12 & 36.39 & 29.15 & 45.92 & 19.17 & 23.08 & 31.94 & 20.09 \\
Llava-llama3-8B & 25.25 & 13.20 & 23.43 & 23.97 & 50.14 & 33.44 & 31.66 & 47.96 & 19.17 & 17.58 & 24.71 & 21.48 \\
Llava-Next-Llama3-8B & 17.70 & 14.80 & 24.57 & 24.48 & 53.74 & 34.43 & 37.69 & 43.88 & 22.63 & 21.98 & 29.66 & 17.55 \\
CogVLM-Chat-7B & 21.91 & 18.40 & 21.43 & 25.77 & 57.06 & 22.95 & 27.14 & 50.00 & 31.87 & 29.67 & 30.42 & 31.87 \\
ShareGPT-4v-7B & 17.70 & 24.00 & 13.71 & 18.56 & 45.98 & 14.10 & 29.15 & 6.12 & 29.10 & 29.12 & 19.39 & 22.63 \\
Qwen2.5VL-3B-Instruct & 22.03 & 16.40 & 24.86 & 27.06 & 57.89 & 23.93 & 55.78 & 38.78 & 18.48 & 24.18 & 26.62 & 18.94 \\
Qwen2.5VL-7B-Instruct & 21.29 & 22.80 & 24.86 & 23.71 & 54.29 & 23.61 & \textbf{63.32} & 46.94 & 34.18 & 32.97 & 24.33 & 24.94 \\
Qwen2.5VL-72B-Instruct & 22.40 & 21.60 & 32.29 & 23.20 & 56.79 & 37.70 & 54.27 & 41.84 & 21.02 & 24.73 & 27.38 & 21.94 \\
Janus-Pro-7B & 22.90 & 20.40 & 26.29 & 28.35 & 57.89 & 36.07 & 43.22 & 50.00 & 32.33 & 26.92 & \underline{36.12} & 27.94 \\
InternVL2.5-8B & 19.06 & 19.20 & 28.00 & 25.77 & 60.39 & 36.07 & 46.73 & 43.88 & 40.42 & 34.07 & 33.84 & 31.64 \\
InternVL2.5-38B & 23.51 & 17.20 & 30.86 & 24.74 & 59.83 & 33.11 & 51.76 & 45.92 & 38.57 & 31.32 & 31.94 & 32.10 \\
QvQ-72B & 33.42 & 22.80 & 27.14 & 30.41 & 54.57 & 44.26 & 54.77 & 52.04 & 34.41 & 31.32 & 26.62 & 31.18 \\
\rowcolor{green!20}
\multicolumn{13}{c}{\textbf{Open-Source Medical-Domain MLLMs}} \\ 
MedDr-80B & 42.95 & 17.60 & 26.57 & 29.38 & 58.17 & \underline{56.72} & 44.72 & 56.12 & 33.03 & 34.62 & 27.38 & 31.87 \\
Llava-Med-7B & 12.75 & \underline{25.60} & 17.43 & 17.78 & 16.90 & 34.43 & 29.15 & 17.35 & 25.64 & 23.08 & 23.57 & 26.56 \\
HuatuoGPT-Vision-7B & 41.46 & 24.40 & 29.43 & 32.47 & 52.36 & 46.89 & 44.72 & 59.18 & 32.56 & 35.71 & 28.52 & 32.79 \\
HuatuoGPT-Vision-34B & 39.48 & 20.40 & 32.57 & 38.66 & 57.34 & 52.79 & 59.30 & 59.18 & 39.49 & \underline{37.91} & 35.36 & 36.03 \\
ColonGPT & 18.44 & \textbf{57.60} & 11.14 & 0.26 & 0.83 & 2.62 & 16.08 & 15.31 & 24.71 & 34.07 & 4.56 & 22.86 \\
\rowcolor{pink!20}
\multicolumn{13}{c}{\textbf{Proprietary MLLMs}} \\ 
Deepseek-V3 & 33.79 & 17.60 & 24.00 & 31.44 & 60.38 & 42.30 & 55.28 & 53.06 & 31.18 & 34.07 & 27.38 & 26.79 \\
Grok-3 & 39.11 & 16.00 & 28.29 & 26.55 & 58.73 & 38.03 & 50.75 & 60.20 & \underline{41.57} & 32.97 & 27.00 & 37.88 \\
Claude-3.7-Sonnet & 43.07 & 13.20 & 31.43 & 24.48 & 55.68 & 51.48 & 50.25 & 44.90 & 30.48 & 35.71 & 30.04 & 36.26 \\
GPT-4o & \textbf{49.38} & 19.20 & \underline{36.00} & \underline{45.88} & \underline{64.54} & 52.79 & \underline{62.81} & \textbf{65.31} & 32.79 & 35.71 & 34.98 & \underline{43.65} \\
Gemini-2.5-Pro & \underline{49.01} & 19.20 & \textbf{44.29} & \textbf{47.94} & \textbf{65.10} & \textbf{71.15} & \underline{62.81} & \underline{63.27} & \textbf{49.19} & \textbf{38.46} & \textbf{51.33} & \textbf{50.58} \\
\bottomrule
\end{tabular}
\end{adjustbox}
\end{table}

\subsection{Case Study}
In this section, we conduct a case study analysis of multiple MLLMs in EndoBench under various scenarios.

\begin{figure}[p]
  \centering
  \includegraphics[width=0.72\linewidth]{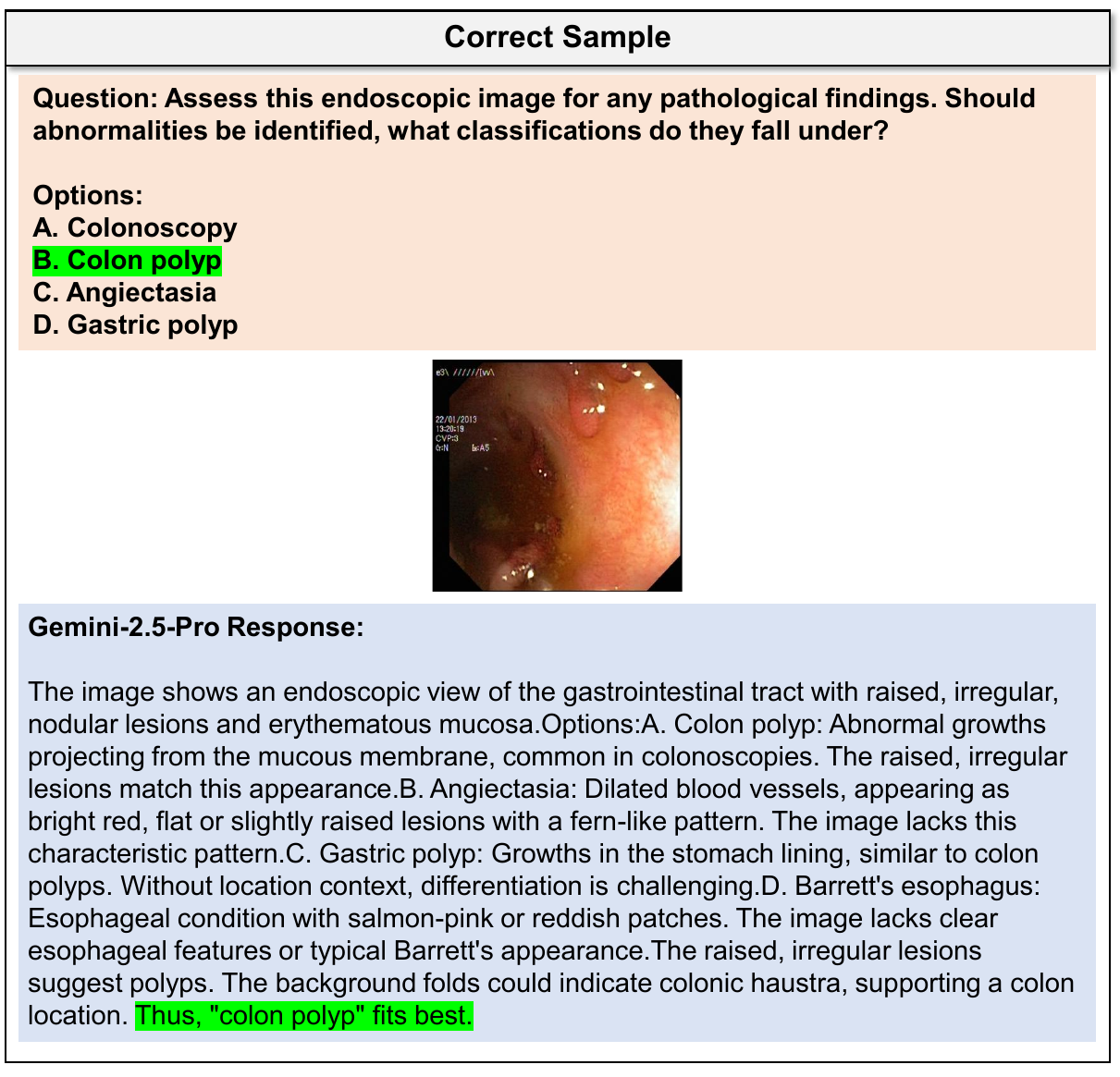}
  \caption{Correct sample}
  \label{fig:case_study_01}
\end{figure}

\begin{figure}[p]
  \centering
  \includegraphics[width=0.72\linewidth]{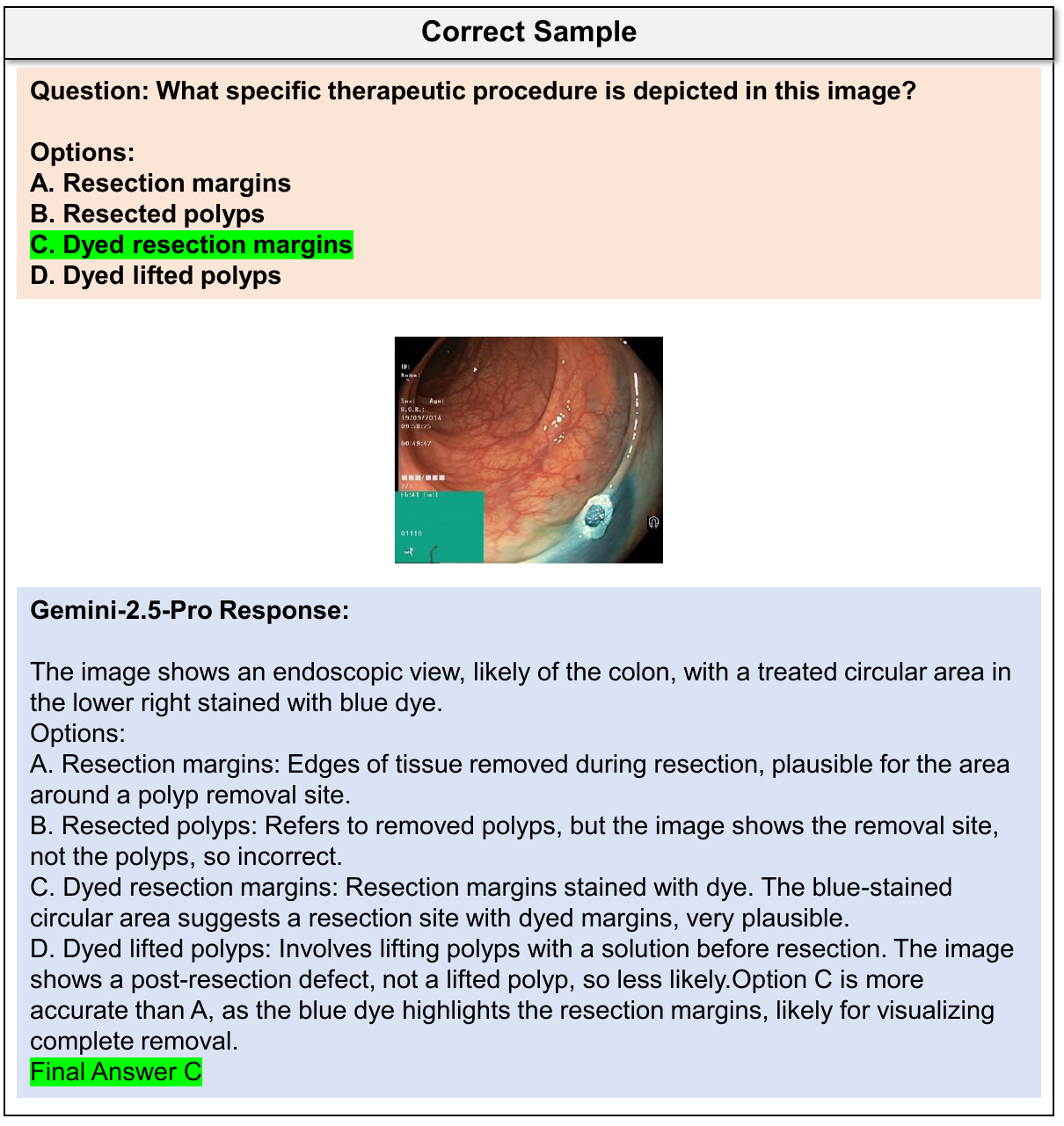}
  \caption{Correct sample}
  \label{fig:case_study_02}
\end{figure}

\begin{figure}[p]
  \centering
  \includegraphics[width=0.72\linewidth]{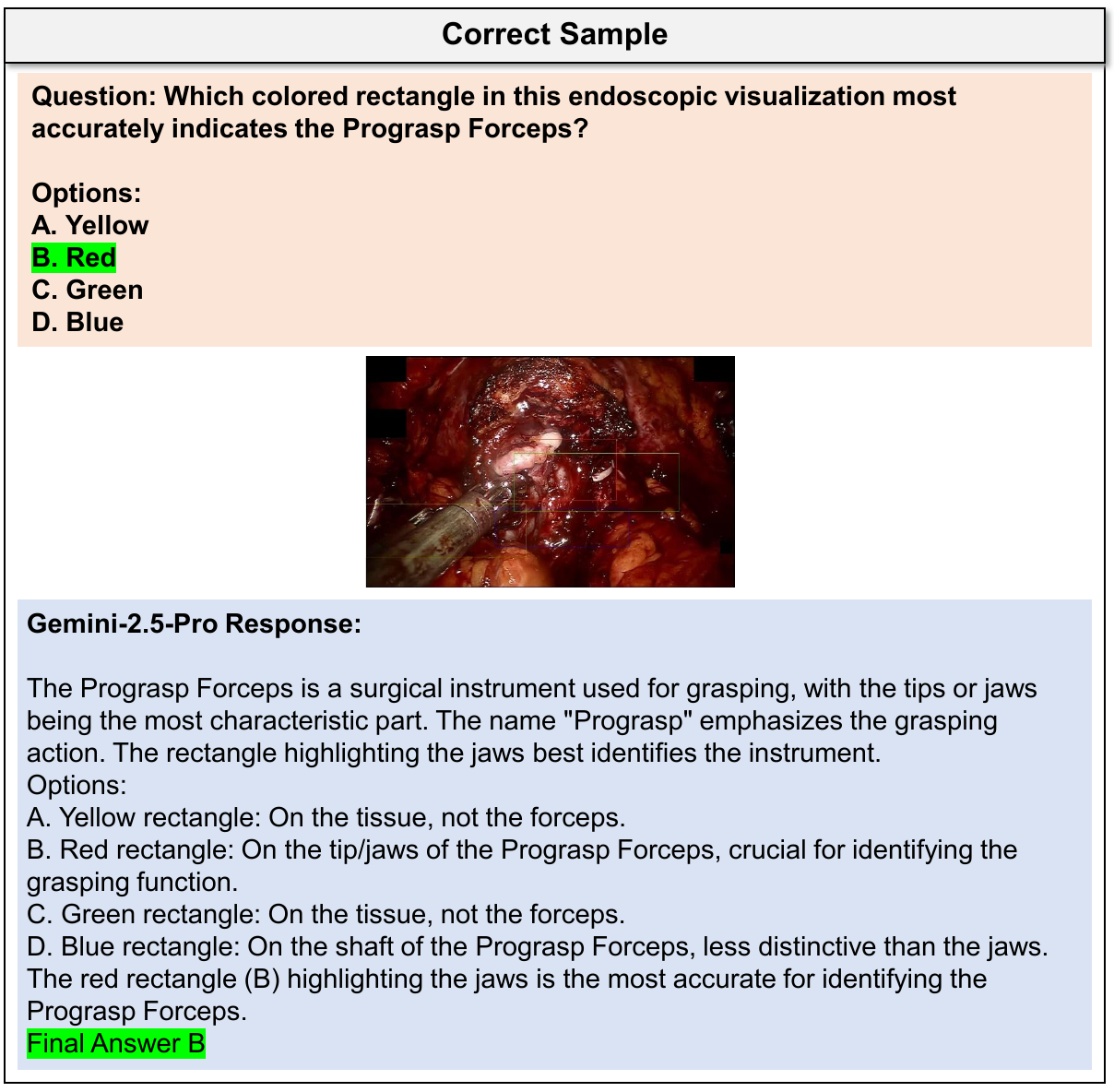}
  \caption{Correct sample}
  \label{fig:case_study_03}
\end{figure}

\begin{figure}[p]
  \centering
  \includegraphics[width=0.72\linewidth]{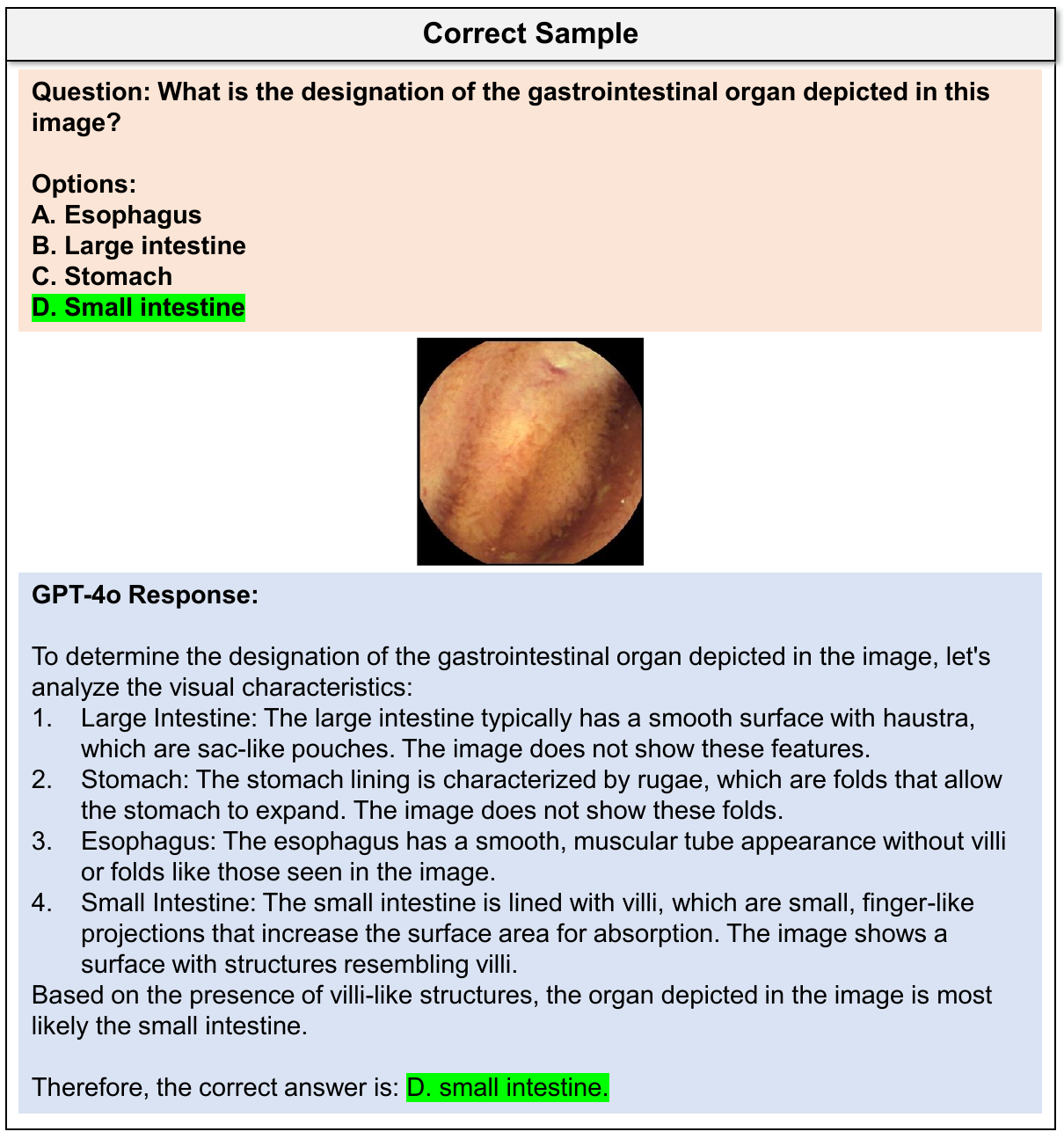}
  \caption{Correct sample}
  \label{fig:case_study_04}
\end{figure}

\begin{figure}[p]
  \centering
  \includegraphics[width=0.72\linewidth]{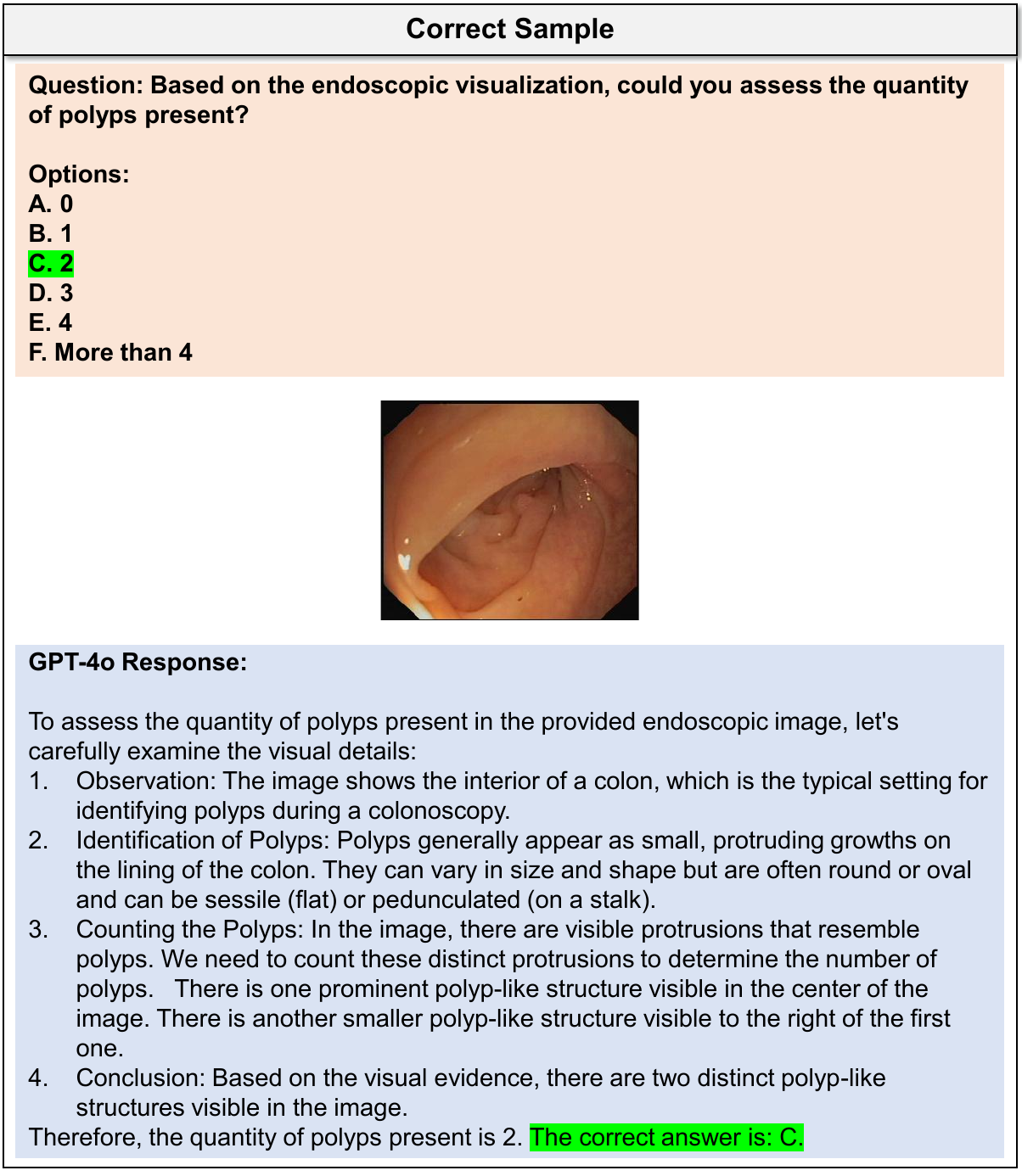}
  \caption{Correct sample}
  \label{fig:case_study_05}
\end{figure}

\begin{figure}[p]
  \centering
  \includegraphics[width=0.72\linewidth]{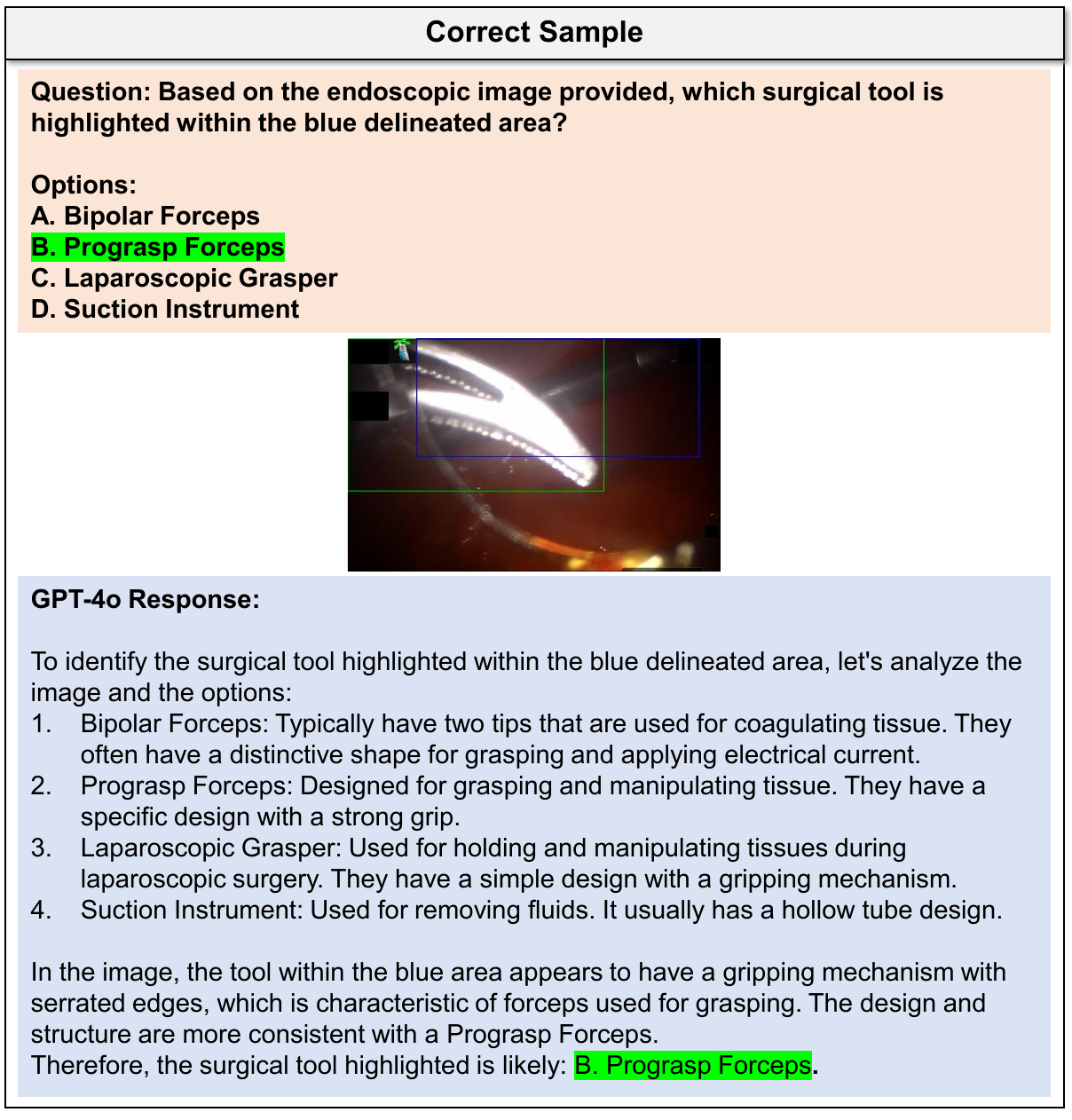}
  \caption{Correct sample}
  \label{fig:case_study_06}
\end{figure}

\textbf{Correct Samples.} These figures (Fig.~\ref{fig:case_study_01}-\ref{fig:case_study_06}) highlight exemplary performances by leading models such as Gemini-2.5-Pro \cite{team2023gemini} and GPT-4o \cite{GPT4o}. These models demonstrate robust capabilities in accurately interpreting endoscopic images and providing clinically relevant responses, highlighting their potential for assisting in real-world endoscopic analysis.

\textbf{Error Analysis.} The errors observed in the case studies are categorized into four types, each highlighting distinct limitations in the performance of multimodal large language models in medical applications.

\begin{figure}[p]
  \centering
  \includegraphics[width=0.72\linewidth]{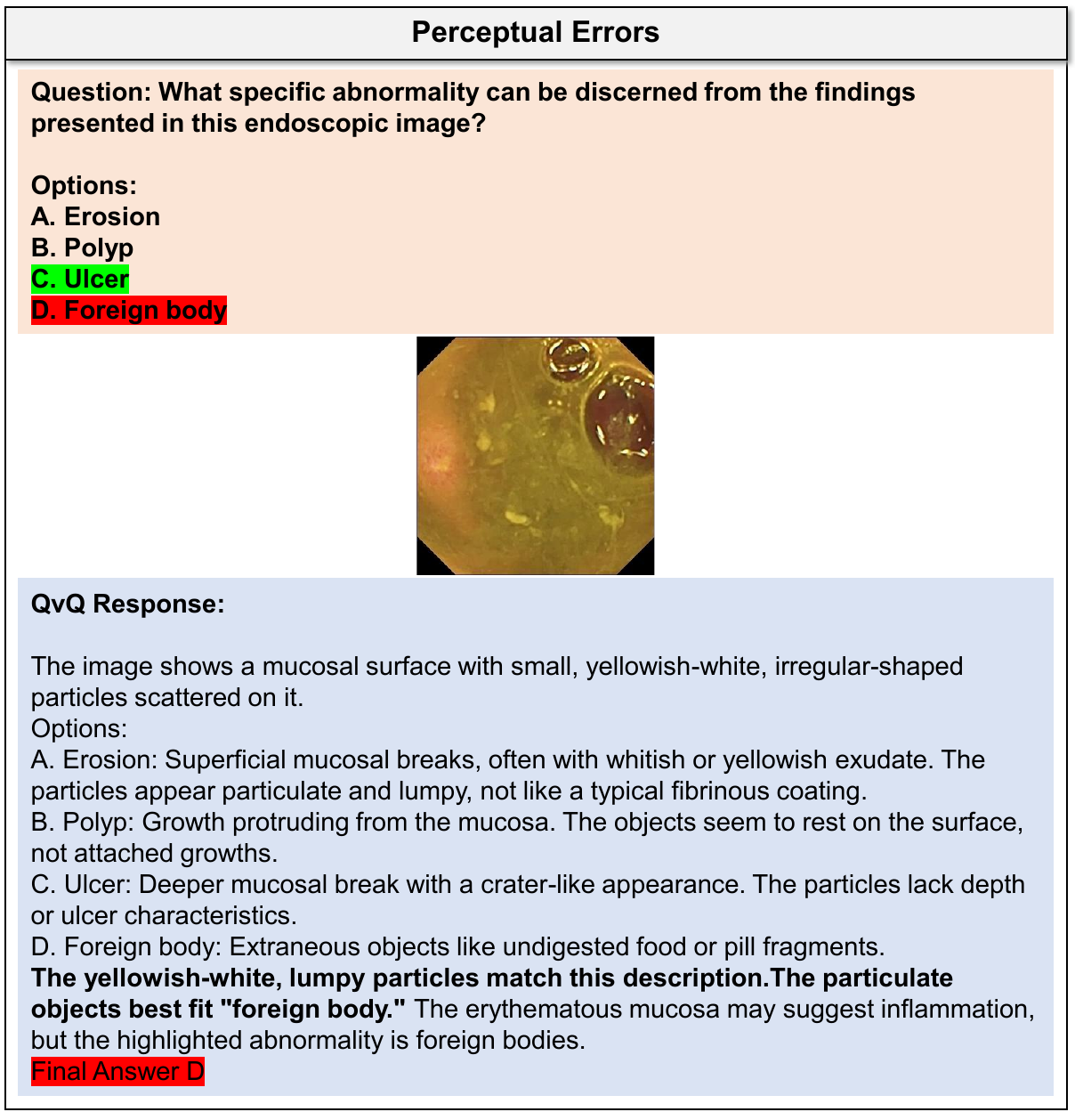}
  \caption{Error sample}
  \label{fig:case_study_07}
\end{figure}

\begin{figure}[p]
  \centering
  \includegraphics[width=0.72\linewidth]{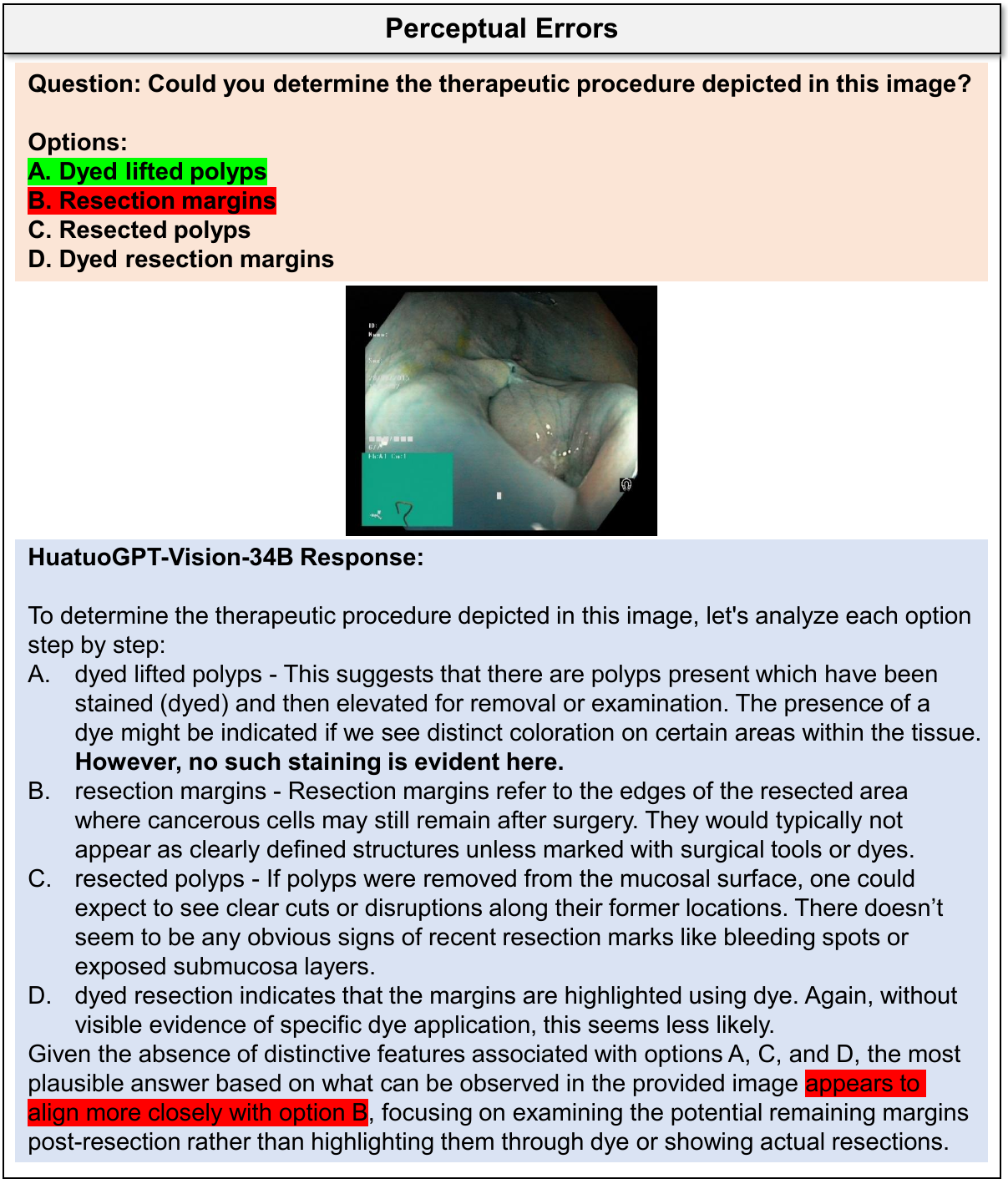}
  \caption{Error sample}
  \label{fig:case_study_08}
\end{figure}

\begin{figure}[p]
  \centering
  \includegraphics[width=0.72\linewidth]{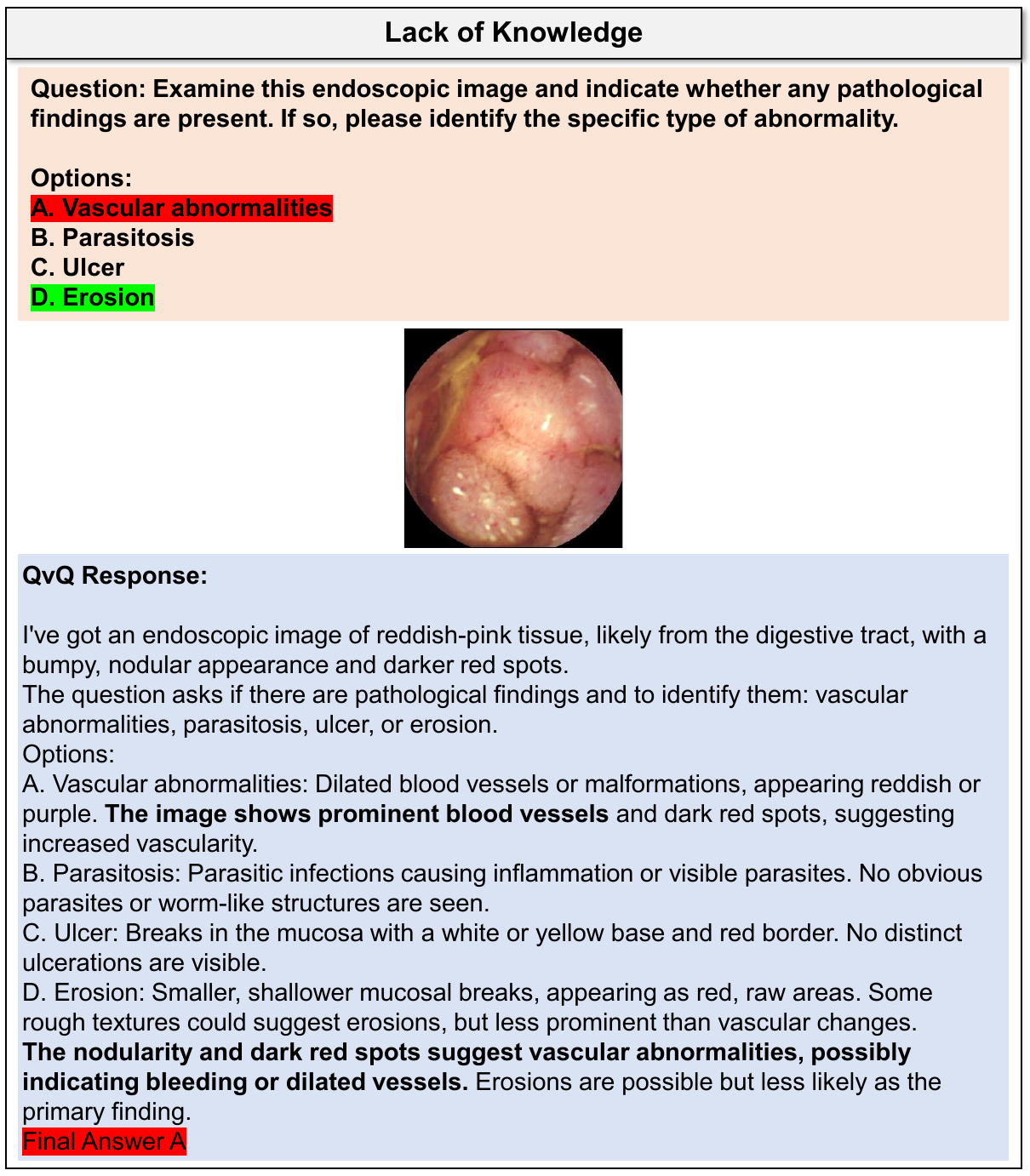}
  \caption{Error sample}
  \label{fig:case_study_09}
\end{figure}

\begin{figure}[p]
  \centering
  \includegraphics[width=0.72\linewidth]{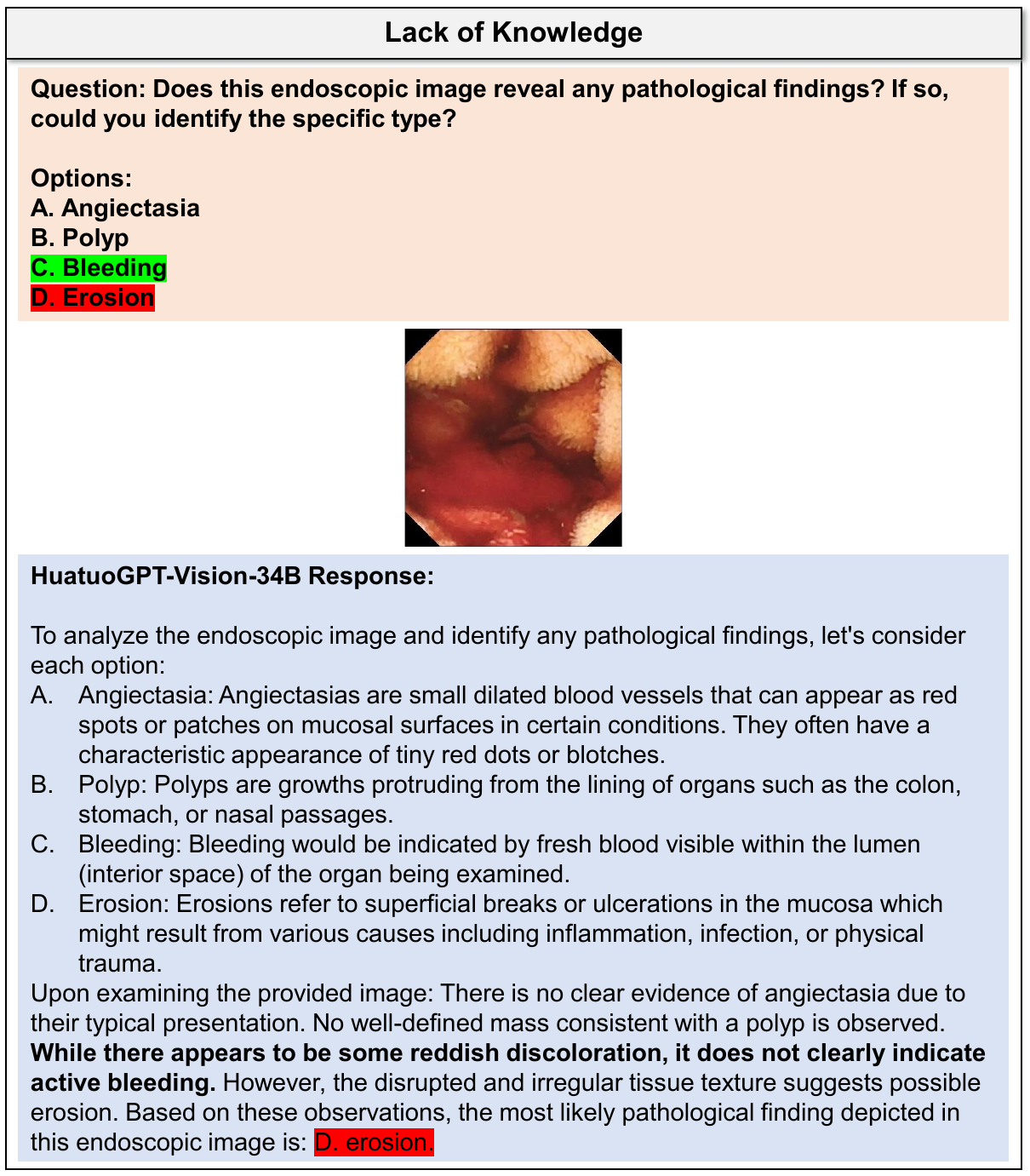}
  \caption{Error sample}
  \label{fig:case_study_10}
\end{figure}

\begin{figure}[p]
  \centering
  \includegraphics[width=0.72\linewidth]{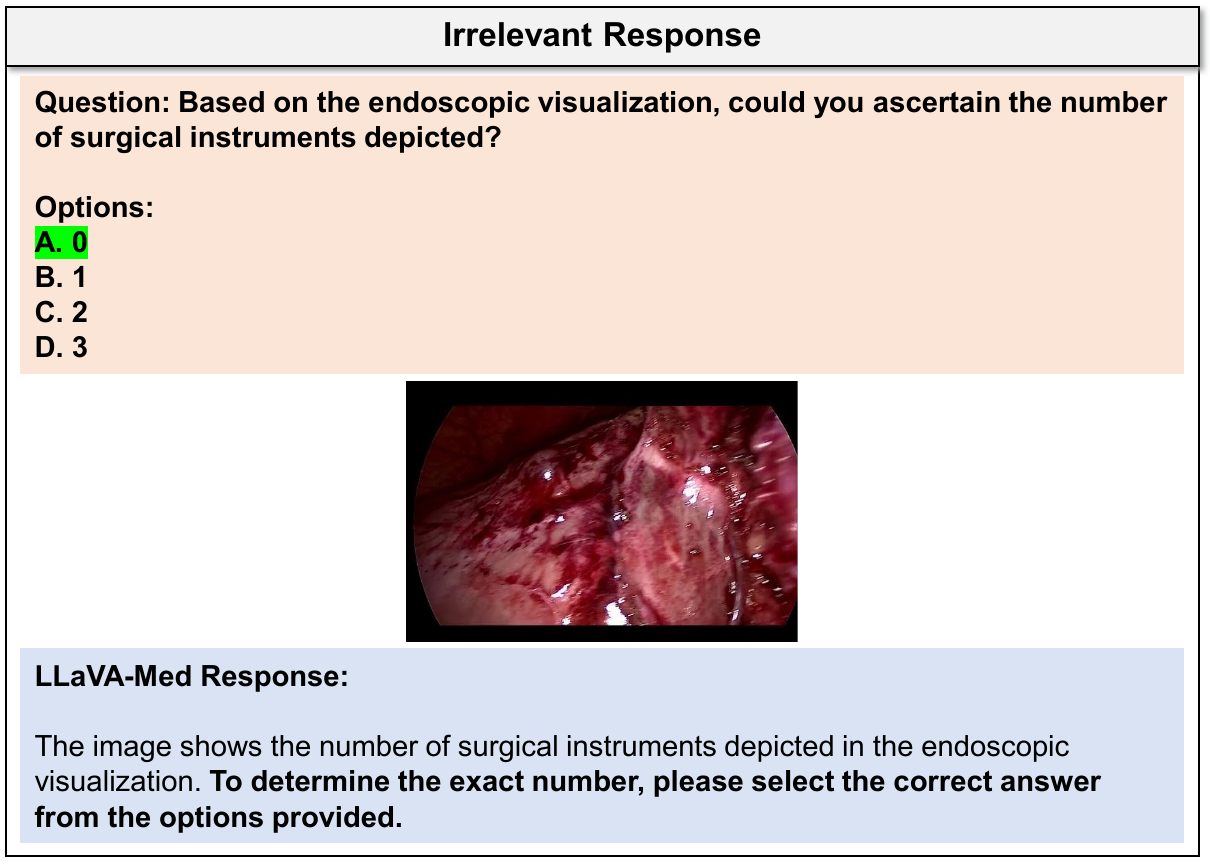}
  \caption{Error sample}
  \label{fig:case_study_11}
\end{figure}

\begin{figure}[p]
  \centering
  \includegraphics[width=0.72\linewidth]{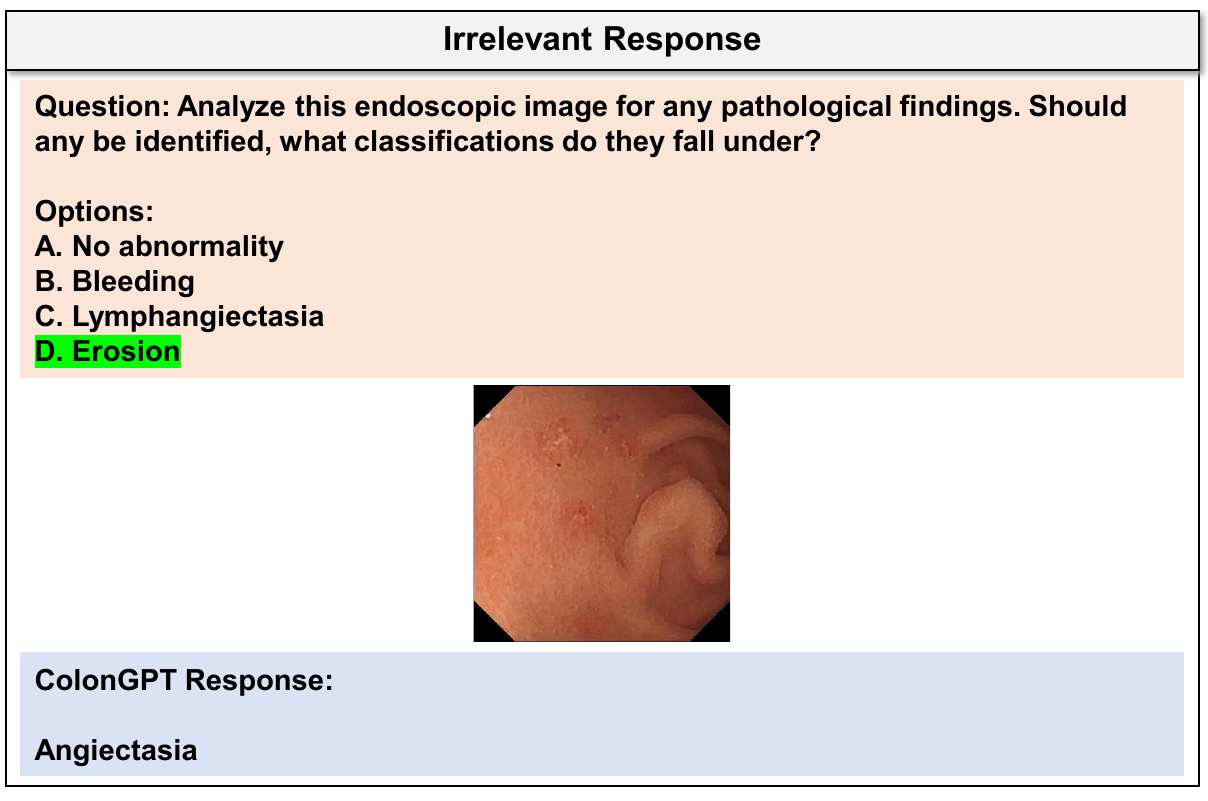}
  \caption{Error sample}
  \label{fig:case_study_12}
\end{figure}

\begin{figure}[p]
  \centering
  \includegraphics[width=0.72\linewidth]{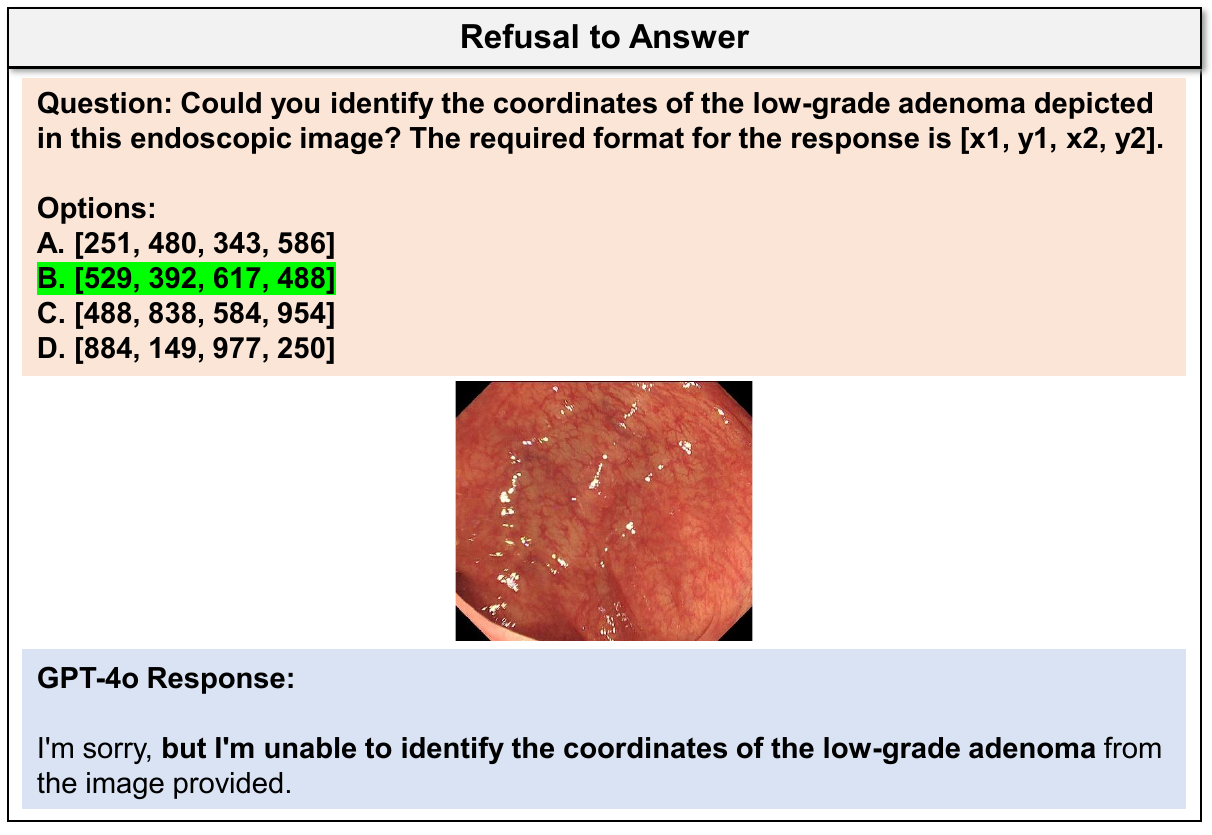}
  \caption{Error sample}
  \label{fig:case_study_13}
\end{figure}

\begin{figure}[p]
  \centering
  \includegraphics[width=0.72\linewidth]{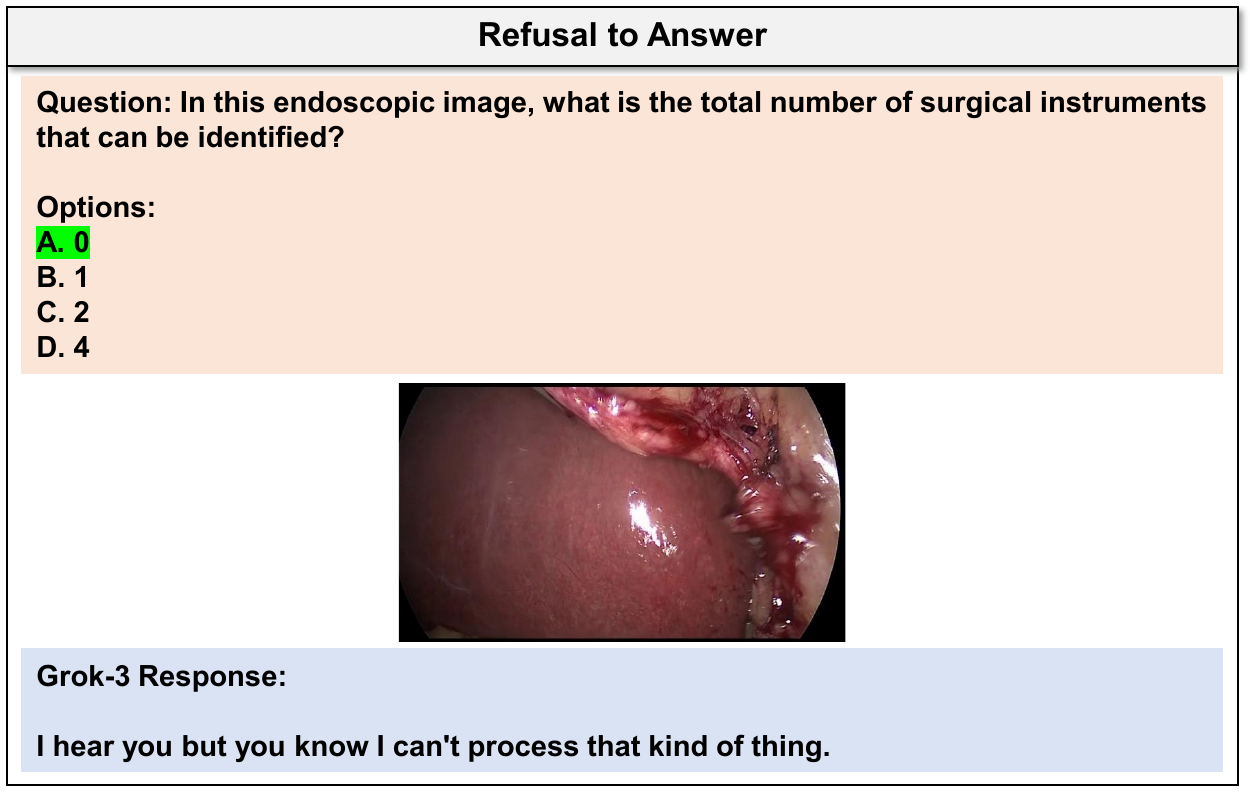}
  \caption{Error sample}
  \label{fig:case_study_14}
\end{figure}

\begin{itemize}[leftmargin=0.5cm]
\item \textbf{Perceptual Errors:} MLLMs may struggle to accurately perceive or interpret visual information in images, including failing to detect critical objects, misidentifying elements, or overlooking essential details. In Fig. \ref{fig:case_study_07}, QvQ-72B~\cite{qvq} fails to recognize erythematous areas and focuses on irrelevant yellow-white granules. Similarly, in Fig. \ref{fig:case_study_08}, HuatuoGPT-Vision-34B~\cite{huatuogptVision} overlooks that the mucosa has been stained blue, leading to an incorrect interpretation of the scene. These indicate a limitation in the model’s ability to accurately recognize clinically significant visual patterns.
\item \textbf{Lack of Knowledge:} MLLMs may accurately identify visual elements in an image and comprehend the question but still provide incorrect answers due to insufficient medical domain expertise. This manifests as misinterpretations of clinical signs or failure to differentiate between similar medical conditions. For instance, in Fig. \ref{fig:case_study_09}, QvQ-72B~\cite{qvq} correctly identifies low-level visual features, such as red points in the image, but misinterprets them as blood vessels. Similarly, in Fig. \ref{fig:case_study_10}, HuatuoGPT-Vision-34B~\cite{huatuogptVision} notices prominent bright red areas in the image during reasoning but fails to interpret them as bleeding, leading to an inaccurate response. These errors highlight a deficiency in domain-specific medical knowledge, where the model fails to contextualize visual cues with appropriate clinical understanding.
\item \textbf{Irrelevant Response:} MLLMs sometimes generate responses that are unrelated to the user’s query, producing irrelevant, incomplete, or incomprehensible information that fails to address  the question. For example, in Fig. \ref{fig:case_study_11}, LLaVA-Med \cite{Llava-med} is asked to determine the number of surgical instruments in an endoscopic image but outputs a tautological restatement of the query, lacking any clinical insight. In another case, Fig. \ref{fig:case_study_12}, ColonGPT \cite{colongpt} is tasked with classifying pathological findings in an endoscopic image but outputs a term unrelated to the provided options and observed pathology. These case studies emphasize the need for improved medical knowledge integration and enhanced perceptual capabilities to bridge the gap between current MLLM performance and clinical requirements.
\item \textbf{Refusal to Answer:} Certain MLLMs, particularly proprietary ones, are designed to decline responses to questions involving sensitive information, ethical dilemmas, or requiring professional medical advice to ensure safety and compliance. For example, in Fig. \ref{fig:case_study_13}, GPT-4o \cite{GPT4o} is asked to identify the coordinates of a low-grade adenoma in an endoscopic image but states it is unable to provide the coordinates. Likewise, in Fig. \ref{fig:case_study_14}, Grok-3 \cite{grok3} is tasked with counting surgical instruments in an endoscopic image but explicitly refuses, citing its inability to process such requests. These cases highlight the need for enhanced technical capabilities and clearer ethical guidelines to balance safety with clinical utility in MLLM responses.
\end{itemize}

\section{Limitations}
While our current work provides a comprehensive benchmark for 2D endoscopic image analysis, this approach has inherent limitations in clinical applicability. Static 2D images cannot capture critical spatial-depth relationships (e.g., polyp morphology assessment) or temporal dynamics (e.g., bleeding source localization), which are essential for accurate diagnosis and surgical planning. These constraints highlight the need to evolve toward 3D endoscopic video analysis, where volumetric reconstruction and motion context could enable transformative applications like real-time surgical navigation, instrument tracking, and dynamic lesion characterization. Future research must address the computational and annotation challenges of 3D video to achieve clinically viable systems that complement physician decision-making in complex endoscopic procedures.

Moreover, our work still involves a key remaining challenge of integrating AI into complete clinical workflows, particularly when handling ambiguous cases requiring expert physician interpretation. These complex scenarios demand additional rigorous clinical validation and nuanced clinical judgment that current systems cannot fully replicate. Future work should address these limitations through expanded validation studies and improved algorithmic handling of diagnostic uncertainties. The path to reliable clinical implementation requires both technological advances in AI interpretation and careful workflow integration to complement - rather than replace - physician expertise in challenging cases.

Furthermore, to enable reliable AI-assisted endoscopic diagnosis, future research should pursue: (1) multicenter clinical trials to validate performance across diverse populations and settings; (2) development of standardized benchmarks assessing diagnostic accuracy, workflow integration, and clinical utility; and (3) establishment of ethical frameworks addressing data privacy, algorithmic bias, and physician-AI collaboration. These efforts must focus particularly on challenging areas like indeterminate cases requiring human expertise. Only through such rigorous validation and standardization can we ensure these technologies meet clinical needs while maintaining patient safety and upholding ethical standards in medical practice.

\section{Potential Improvement Methodology}
Based on the evaluations of EndoBench, existing MLLMs still have a long way to go before they can be applied clinically. To bridge this performance gap, several key methodologies can be explored, ranging from training-free prompting strategies to more intensive model adaptation techniques.

First, advanced prompt engineering offers a direct path to enhancing performance without additional model training. As shown in Table \ref{tab:model-performance}, our extended evaluation of Zero-Shot Chain-of-Thought (CoT) prompting \cite{CoT} reveals its model-dependent efficacy. While it improved accuracy for robust models like GPT-4o and Gemini-2.5-Pro, it degraded the performance of HuatuoGPT-Vision. This suggests that compelling a model to articulate a reasoning path can paradoxically increase the risk of hallucination when its internal knowledge is not sufficiently grounded \cite{cot_use01,cot_use02,cot_use03}. More robust prompting techniques, such as Self-Consistency (SC) \cite{Self-Consistency} and Self-Refine \cite{Self-refine}, which leverage multiple reasoning paths and iterative feedback, or domain-adapted strategies like MedPrompts \cite{medprompt}, may offer more consistent improvements for complex zero-shot clinical reasoning.

\begin{table}[htbp]
\centering
\caption{Performance comparison of different models with Direct Inference vs. Zero-shot CoT.}
\label{tab:model-performance}
\resizebox{\textwidth}{!}{
\begin{tabular}{lccccc}
\toprule
\textbf{Model} & \textbf{Qwen2.5-VL-7B} & \textbf{HuatuoGPT-Vision-7B} & \textbf{HuatuoGPT-Vision-34B} & \textbf{GPT-4o} & \textbf{Gemini-2.5-Pro} \\
\midrule
Direct inference & 27.63 & \textbf{35.57} & \textbf{39.58} & 41.69 & 49.53 \\
Zero-shot CoT    & \textbf{32.35} & 28.53          & 32.14          & \textbf{42.11} & \textbf{61.67} \\
\bottomrule
\end{tabular}%
}
\end{table}

Second, as highlighted in our findings (Observation 2), domain-specific Supervised Fine-Tuning (SFT) represents a powerful method for instilling specialized knowledge. Future work should prioritize fine-tuning MLLMs on high-quality, curated medical instruction datasets. Such datasets could comprise clinical dialogues, endoscopic procedural reports, and synthetic question-answer pairs tailored to endoscopic analysis. This process is crucial for aligning the model’s internal representations and reasoning behavior with the specific nuances and demands of the clinical domain.

Finally, to bolster factual accuracy and mitigate hallucinations, integrating structured medical knowledge is another critical direction. This can be achieved through techniques like Retrieval-Augmented Generation (RAG) \cite{medgraphrag, MMed-RAG} or by fusing embeddings from clinical knowledge graphs \cite{lopez2025clinical}. By dynamically providing the model with contextual support from verified medical sources during inference, these methods can significantly reduce factual inaccuracies and enhance the reliability of generated outputs.

While a comprehensive implementation of these approaches extends beyond the scope of this benchmark, they collectively outline a clear roadmap for future research. Pursuing these directions is essential for developing MLLMs that are safe, effective, and clinically viable.

\section{Potential Negative Social Impacts}

We propose a comprehensive benchmark for MLLMs in endoscopy by integrating twenty public endoscopy datasets, which cover diverse endoscopic image types and clinical terminology. Regarding social implications, potential negative impacts may include:

\begin{itemize}[leftmargin=0.5cm]
    \item \textit{Diagnostic Inequality Risks.} The anonymization of public datasets, while protecting patient privacy, results in the loss of critical demographic information (e.g., area, ethnicity). This data gap may lead to biased AI models performing unequally across different population groups, potentially exacerbating healthcare disparities. For instance, the model might achieve higher diagnostic accuracy for specific demographic groups while underperforming for others, particularly underrepresented populations.
    \item \textit{Technological Exclusion of Underserved Healthcare Systems.} The current benchmark's focus on advanced endoscopic modalities, including wireless capsule endoscopy, risks marginalizing medical institutions in low-resource settings that lack access to such technologies. This creates an infrastructural bias in AI development, where models are optimized primarily for well-equipped hospitals while failing to address the diagnostic needs of resource-constrained clinics. Consequently, the benefits of AI-assisted endoscopy may disproportionately favor high-income regions, exacerbating global healthcare inequities. To promote inclusive progress, future benchmarks should evaluate model performance across diverse clinical environments—from basic to advanced endoscopic systems—ensuring these tools remain accessible and effective regardless of a facility's technological capacity.
    \item \textit{Security Vulnerabilities in MLLMs Diagnostics.} Current benchmarks exhibit security vulnerabilities by failing to reject harmful inputs, potentially generating dangerous misdiagnoses. These systems lack robust safeguards against adversarial attacks, enabling malicious actors to induce false medical outputs. Such flaws could compromise patient safety through weaponized misinformation or engineered diagnostic errors. Addressing these risks requires specialized medical guardrails and rigorous adversarial testing for clinical deployment.
    \item \textit{Ethical Implications.} Deploying MLLMs in endoscopic scenarios raises ethical concerns such as patient privacy, model transparency, and accountability, which could lead to misdiagnosis or harm. Thus, developing robust safeguards and clear responsibility frameworks is necessary. Specifically, data collection must have approval from the respective institutional review boards, with newly added data de-identified to remove all patient-related information. For private databases prohibiting direct data use with external APIs like OpenAI, more capable models such as ChatGPT-o1 can be tested via APIs through HIPAA-compliant cloud providers like Azure. The medical benchmark must undergo rigorous clinical review by experienced physicians.
\end{itemize}

\clearpage 

\clearpage

\clearpage

\clearpage

\clearpage

\clearpage

\clearpage 

\newpage

\end{document}